\documentclass[journal]{IEEEtran}
\usepackage{amsmath,amssymb,amsfonts}
\usepackage{bm}
\usepackage{booktabs}
\usepackage{graphicx}
\usepackage{cite}
\usepackage{stfloats}
\usepackage{tikz}
\usetikzlibrary{backgrounds}
\usetikzlibrary{arrows.meta,positioning,calc,matrix,fit}
\usepackage{multirow}
\usepackage{xcolor}
\usepackage{colortbl}
\pdftrailerid{}
\usepackage{hyperref}
\hypersetup{hidelinks,pdfauthor={},pdftitle={},pdfsubject={},pdfkeywords={},pdfcreator={},pdfproducer={}}

\definecolor{posblue}{RGB}{0,114,178}
\definecolor{negred}{RGB}{213,94,0}
\definecolor{twitchorange}{RGB}{230,159,0}
\definecolor{signalgreen}{RGB}{0,158,115}
\definecolor{jointpurple}{RGB}{204,121,167}
\colorlet{best}{signalgreen!35}
\colorlet{worst}{negred!30}

\begin{document}

\title{Motoneuron-Inspired Sampling \\ for Model Predictive Path Integral Control}

% Switch to \blindfalse only after acceptance, for the final author version.
\newif\ifblind
\blindfalse
\ifblind
  % Phantom reserves the byline width without printing identifying text.
  \author{\phantom{Alexis~Poignant and Jan~Babi\v{c}}}
  % Do not place \url, \href, or \thanks containing identity-revealing links
  % inside \phantom: invisible link annotations/footnote marks can persist.
\else
  \author{Alexis~Poignant and Jan~Babi\v{c}%
    \thanks{Source code and supplementary animated visualizations are available at
    \url{https://github.com/fleurssauvages/spikeMPPI} and
    \url{https://github.com/fleurssauvages/Spike-MPPI-Manipulation}.}}
\fi

\maketitle

% Declare Fig. 1 immediately after the title, as in the ieeeconf source.
% First-page placement is not guaranteed by IEEEtran.
\begin{figure*}[b]
    \centering
    \includegraphics[width=\textwidth]
    {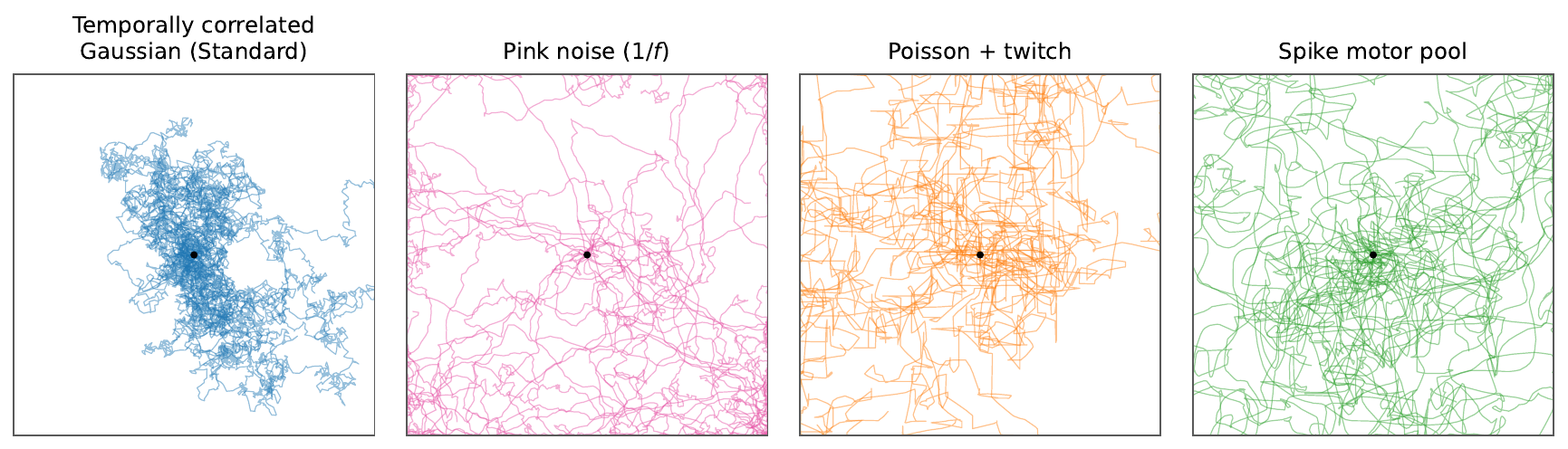}
    \caption{
        Exploration trajectories induced by four sampling
        distributions in a bounded two-dimensional integrator.
        Each panel shows 18 independent trajectories starting
        from the origin (black dot), using identical simulation
        duration and nominal exploration scale.
        Temporally correlated Gaussian sampling (Standard)
        produces predominantly local exploration, whereas
        pink noise generates longer, more coherent excursions.
        Poisson sampling exhibits event-driven, twitch-filtered
        transitions, while Spike sampling produces more structured
        trajectories.
    }
    \label{fig:noise-exploration}
\end{figure*}

\begin{abstract}
Model Predictive Path Integral (MPPI) control relies on stochastic trajectory sampling, and its performance under limited rollout budgets depends strongly on the structure of the proposal distribution. Conventional Gaussian perturbations, even when temporally correlated, transmit high-frequency jitter to the executed commands when only a few rollouts are available. We introduce Spike-MPPI, a motoneuron-inspired proposal in which a common stochastic drive recruits pools of motor units whose discharges are filtered by twitch kernels, yielding sparse, temporally coherent perturbations. We evaluate Spike-MPPI on legged locomotion and manipulation tasks with a Franka FR3 arm in simulation, as well as on a physical robot. Spike consistently produces the smoothest executed commands across all tasks, including against spectrum-matched Gaussian proposals, and a planar ablation identifies the common drive as the largest contributor to this smoothness among the tested motor-pool mechanisms. In manipulation, this smoothness is also accompanied by improved task performance and success, such as in ball balancing, where Gaussian, pink-noise, and spectrum-matched proposals largely fail. On hardware, Spike also yields the shortest task completion times and the smoothest commands. These results show that, beyond its power spectrum, the statistical structure of the sampling is a key design factor for finite-budget MPPI control.
\end{abstract}


\section{Introduction}
\label{sec:introduction}

Model Predictive Path Integral (MPPI) control is a sampling-based stochastic optimal-control method derived from path-integral formulations of optimal control \cite{theodorou2010generalized}. In its receding-horizon form, MPPI evaluates a population of control sequences through the system dynamics and updates the nominal sequence by exponentially weighting low-cost rollouts \cite{williams2015model, williams2017model}. The method is attractive for nonlinear and contact-rich systems because it does not require derivatives of the dynamics or cost function, its rollouts can be evaluated independently in parallel, and it has been demonstrated in demanding real-time control problems such as aggressive autonomous driving \cite{williams2016aggressive, williams2018information}. Under a finite rollout budget, however, performance depends strongly on the proposal distribution from which candidate control trajectories are generated.

This dependence becomes particularly important for direct control of articulated systems. Sampling an independent perturbation for every actuator at every prediction step produces a search space whose dimension grows with both the control horizon and the number of actuators. Several approaches therefore reduce the effective search dimension through temporal or structural parameterizations. Whole-body and spline-interpolated MPPI sample sparse knot or control points and reconstruct the horizon by interpolation \cite{alvarez2025real,miura2024spline}; Model Tensor Planning builds globally diverse candidates from structured graph samples \cite{le2025model}; and single-instance sampling with a dynamic horizon enables efficient real-time task-space MPPI for manipulators \cite{kim2025single}. These developments illustrate how reducing the dimensionality and modifying the structure of candidate trajectories can improve sampling-based control.

The temporal statistics of the proposal distribution constitute another important design choice. Vlahov et al.\ \cite{vlahov2024low} showed that replacing uncorrelated Gaussian perturbations with colored low-frequency noise can produce smoother, more exploratory candidates (Fig.~\ref{fig:noise-exploration}, blue and pink) and equal or better MPPI performance across systems with different actuator response speeds. In related sampling-based optimization, iCEM \cite{pinneri2021sample} demonstrated that temporally correlated actions combined with sample reuse can reduce the number of trajectories required for effective planning. Adaptive importance sampling has also been shown to improve MPPI performance under limited sampling budgets, particularly as action-space dimensionality increases \cite{asmar2023moppi}. Together, these results motivate proposal distributions that concentrate exploration on temporally coherent control trajectories rather than treating every control coordinate independently.

Smooth control generation has also been addressed through modifications to the optimization formulation itself. Smooth MPPI combines input lifting with an additional action cost to reduce control chattering without external post-processing \cite{kim2022smooth}. Time-Correlated MPPI (TC-MPPI) incorporates action-derivative penalties into the stochastic optimal-control formulation and derives a time-correlated conditional Gaussian sampling distribution \cite{lee2025time}. From a control-as-inference perspective, Watson and Peters \cite{watson2023inferring} demonstrated that Gaussian-process action priors can produce smooth control sequences through sequential posterior policy iteration. In contrast, the present work investigates how the statistical organization of the proposal distribution influences closed-loop behavior while retaining a common MPPI optimization and update procedure.

% Drop-in TikZ figure.
% Requires:
%   \usepackage{xcolor}
%   \usepackage{graphicx}
%   \usepackage{tikz}
%   \usetikzlibrary{arrows.meta,positioning,calc,fit,backgrounds}

\definecolor{posblue}{RGB}{0,114,178}
\definecolor{antorange}{RGB}{213,94,0}
\definecolor{jointpurple}{RGB}{120,45,170}

\begin{figure*}[t]
\centering
\resizebox{\textwidth}{!}{%
\begin{tikzpicture}[
    x=1cm,y=1cm,
    >=Latex,
    font=\small,
    panel/.style={draw=black!65, rounded corners=3pt, line width=0.65pt, fill=white},
    block/.style={draw=black!70, rounded corners=2.5pt, line width=0.65pt, fill=white,
                  align=center, inner sep=2.5pt},
    blueblock/.style={block, draw=posblue},
    orangeblock/.style={block, draw=antorange},
    arrow/.style={->, line width=0.85pt, draw=black!80},
    bluearrow/.style={->, line width=0.9pt, draw=posblue},
    orangearrow/.style={->, line width=0.9pt, draw=antorange},
    signal/.style={line width=1.05pt},
    smallaxis/.style={->, thin, draw=black!80},
]

% ============================================================
% (a) POISSON POINT-PROCESS PROPOSAL
% ============================================================
\node[panel, minimum width=18cm, minimum height=3.5cm, anchor=north west] (pa) at (0,18.30) {};
\node[anchor=west, font=\bfseries\large] at (0.15,17.98) {(a) Poisson point-process proposal};

% Inputs -- deliberately text-only.
\node[blueblock, minimum width=2.55cm, minimum height=0.64cm] (aplus) at (1.55,17.06) {
    {\scriptsize\textcolor{posblue}{Positive event rate}}
};
\node[orangeblock, minimum width=2.55cm, minimum height=0.64cm] (aminus) at (1.55,16.20) {
    {\scriptsize\textcolor{antorange}{Negative event rate}}
};
\node[block, minimum width=2.55cm, minimum height=0.64cm] (aprob) at (1.55,15.35) {
    {\scriptsize Recruitment marks}
};

% Signed marked events
\node[block, minimum width=4.15cm, minimum height=2.28cm] (aevents) at (5.65,16.20) {};
\node[anchor=north, font=\scriptsize, align=center] at ($(aevents.north)+(0,-0.0)$) {
    \textbf{Signed marked events $x_{t,m}$}\\[-0.0mm]
    generated by Poisson process
};
\begin{scope}[shift={(4.20,15.82)}]
    \draw[smallaxis] (0,0) -- (2.90,0);
    \node[text=posblue, font=\scriptsize] at (-0.17,0.42) {$+$};
    \node[text=antorange, font=\scriptsize] at (-0.17,-0.42) {$-$};
    \foreach \x/\h in {0.42/0.38,1.05/0.82,1.72/0.62,2.42/0.90}{\draw[signal,posblue] (\x,0)--(\x,\h);}
    \foreach \x/\h in {0.70/0.55,1.42/0.38,2.10/0.72}{\draw[signal,antorange] (\x,0)--(\x,-\h);}
\end{scope}
\draw[bluearrow] (aplus.east) -- ($(aevents.west)+(0,0.58)$);
\draw[orangearrow] (aminus.east) -- (aevents.west);
\draw[arrow] (aprob.east) -- ($(aevents.west)+(0,-0.58)$);

% Twitch responses
\node[block, minimum width=4.18cm, minimum height=2.28cm] (atwitch) at (10.42,16.20) {};
\node[anchor=north, font=\scriptsize, align=center] at ($(atwitch.north)+(0,-0.12)$) {
    \textbf{Shifted twitch responses}\\[-0.2mm]
    from kernel $\kappa$
};
\begin{scope}[shift={(8.94,15.80)}]
    \draw[smallaxis] (0,0) -- (2.96,0);
    \draw[signal,posblue,dashed]
        (0.12,0) .. controls (0.22,0.02) and (0.34,0.44) .. (0.67,0.49)
        .. controls (1.02,0.49) and (1.12,0.14) .. (1.48,0.03);
    \draw[signal,posblue,dashed]
        (0.80,0) .. controls (0.92,0.03) and (1.05,0.68) .. (1.40,0.72)
        .. controls (1.79,0.72) and (2.00,0.22) .. (2.39,0.04);
    \draw[signal,posblue,dashed]
        (1.58,0) .. controls (1.69,0.02) and (1.80,0.40) .. (2.08,0.43)
        .. controls (2.37,0.43) and (2.55,0.14) .. (2.79,0.03);
    \draw[signal,antorange,dashed]
        (0.50,0) .. controls (0.62,-0.02) and (0.75,-0.52) .. (1.08,-0.58)
        .. controls (1.40,-0.58) and (1.62,-0.18) .. (1.92,-0.03);
    \draw[signal,antorange,dashed]
        (1.40,0) .. controls (1.53,-0.02) and (1.66,-0.43) .. (1.99,-0.47)
        .. controls (2.30,-0.47) and (2.52,-0.16) .. (2.80,-0.03);
\end{scope}
\draw[arrow] (aevents.east) -- (atwitch.west);

% Summed perturbation
\node[block, minimum width=3.55cm, minimum height=2.28cm] (asum) at (15.36,16.20) {};
\node[anchor=north, font=\scriptsize, align=center] at ($(asum.north)+(0,-0.12)$) {
    \textbf{Summed perturbation $z_{t,m}$}\\[-0.2mm]
    for motor channel $m$
};
\begin{scope}[shift={(14.00,15.77)}]
    \draw[smallaxis] (0,0) -- (2.62,0);
    \draw[signal,jointpurple]
      (0.05,0.02)
      .. controls (.20,.04) and (.27,.33) .. (.47,.39)
      .. controls (.72,.46) and (.83,-.27) .. (1.06,-.30)
      .. controls (1.33,-.33) and (1.41,.52) .. (1.70,.46)
      .. controls (1.97,.40) and (2.07,-.18) .. (2.30,-.10)
      .. controls (2.48,-.04) and (2.56,.18) .. (2.60,.07);
\end{scope}
\draw[arrow] (atwitch.east) -- (asum.west);

% ============================================================
% (b) SPIKE MOTOR-POOL DISCHARGE MODEL
% ============================================================
\node[panel, minimum width=18cm, minimum height=7.25cm, anchor=north west] (pb) at (0,14.52) {};
\node[anchor=west, font=\bfseries\large] at (0.15,14.20) {(b) Spike motor-pool discharge model};

% Large colored lanes for clarity.
\begin{scope}[on background layer]
    \fill[posblue!7, rounded corners=3pt] (4.15,11.16) rectangle (17.72,13.70);
    \fill[antorange!7, rounded corners=3pt] (4.15,8.18) rectangle (17.72,10.72);
\end{scope}

% Common stochastic drive
\node[block, minimum width=3.25cm, minimum height=3.36cm] (bdrive) at (1.85,10.94) {};
\node[anchor=north, font=\scriptsize\bfseries, align=center] at ($(bdrive.north)+(0,-0.15)$) {Common stochastic\\drive};
\begin{scope}[shift={(0.72,10.55)}]
    \draw[smallaxis] (0,0) -- (2.20,0);
    \draw[gray!75, line width=.80pt] plot[smooth] coordinates {
        (0.02,.02)(.10,.18)(.18,-.12)(.26,.28)(.34,-.17)(.42,.08)(.50,.32)(.58,-.25)
        (.66,.12)(.74,-.05)(.82,.26)(.90,-.13)(.98,.21)(1.06,-.24)(1.14,.09)(1.22,.30)
        (1.30,-.20)(1.38,.05)(1.46,.19)(1.54,-.10)(1.62,.25)(1.70,-.28)(1.78,.12)(1.86,-.06)
        (1.94,.31)(2.02,-.17)(2.10,.04)(2.18,.20)};
\end{scope}
\node[font=\tiny, align=center, text=black!65] at (1.85,9.55)
    {shared, temporally correlated\\signal};

% Pool split.
\node[blueblock, minimum width=1.35cm, minimum height=1.22cm] (bpoolp) at (4.88,12.42) {
    {\scriptsize\bfseries Agonist}\\[-0.4mm]
    {\scriptsize pool $(+)$}
};
\node[orangeblock, minimum width=1.35cm, minimum height=1.22cm] (bpoolm) at (4.88,9.44) {
    {\scriptsize\bfseries Antagonist}\\[-0.4mm]
    {\scriptsize pool $(-)$}
};
\draw[bluearrow] (bdrive.east) -- ++(.36,0) |- (bpoolp.west);
\draw[orangearrow] (bdrive.east) -- ++(.36,0) |- (bpoolm.west);

% Ordered recruitment and hysteresis: multi-step recruitment and derecruitment stairs.
\node[blueblock, minimum width=4.20cm, minimum height=2.18cm] (brecp) at (8.08,12.42) {};
\node[orangeblock, minimum width=4.20cm, minimum height=2.18cm] (brecm) at (8.08,9.44) {};
\draw[bluearrow] (bpoolp.east) -- (brecp.west);
\draw[orangearrow] (bpoolm.east) -- (brecm.west);
\node[anchor=north, font=\scriptsize, align=center] at ($(brecp.north)+(0,-.09)$)
    {\textbf{Ordered recruitment}\\[-0.6mm]\textbf{with hysteresis}\\[-0.5mm]{\tiny solid: recruit \qquad dashed: derecruit}};
\node[anchor=north, font=\scriptsize, align=center] at ($(brecm.north)+(0,-.09)$)
    {\textbf{Ordered recruitment}\\[-0.6mm]\textbf{with hysteresis}\\[-0.5mm]{\tiny solid: recruit \qquad dashed: derecruit}};

\begin{scope}[shift={(6.55,11.48)}]
    \draw[smallaxis] (0,0)--(2.90,0);
    \node[font=\tiny,anchor=north east] at (2.86,+.3) {drive};
    \draw[smallaxis] (0,0)--(0,.88);
    \node[font=\tiny,rotate=90] at (-.34,.45) {active units};
    % Increasing-drive recruitment staircase.
    \draw[posblue,line width=1.0pt]
      (.10,.08)--(.62,.08)--(.62,.25)--(1.14,.25)--(1.14,.42)--(1.66,.42)--
      (1.66,.59)--(2.18,.59)--(2.18,.76)--(2.78,.76);
    % Decreasing-drive derecruitment staircase, shifted left (lower thresholds).
    \draw[posblue!60,dashed,line width=1.0pt]
      (.10,.08)--(.42,.08)--(.42,.25)--(.94,.25)--(.94,.42)--(1.46,.42)--
      (1.46,.59)--(1.98,.59)--(1.98,.76)--(2.78,.76);
\end{scope}
\begin{scope}[shift={(6.55,8.50)}]
    \draw[smallaxis] (0,0)--(2.90,0);
    \node[font=\tiny,anchor=north east] at (2.86,+.3) {drive};
    \draw[smallaxis] (0,0)--(0,.88);
    \node[font=\tiny,rotate=90] at (-.34,.45) {active units};
    \draw[antorange,line width=1.0pt]
      (.10,.08)--(.62,.08)--(.62,.25)--(1.14,.25)--(1.14,.42)--(1.66,.42)--
      (1.66,.59)--(2.18,.59)--(2.18,.76)--(2.78,.76);
    \draw[antorange!60,dashed,line width=1.0pt]
      (.10,.08)--(.42,.08)--(.42,.25)--(.94,.25)--(.94,.42)--(1.46,.42)--
      (1.46,.59)--(1.98,.59)--(1.98,.76)--(2.78,.76);
\end{scope}

% Rate coding
\node[blueblock, minimum width=2.65cm, minimum height=2.18cm] (bratep) at (12.15,12.42) {};
\node[orangeblock, minimum width=2.65cm, minimum height=2.18cm] (bratem) at (12.15,9.44) {};
\node[anchor=north, font=\scriptsize\bfseries] at ($(bratep.north)+(0,-.12)$) {Rate coding};
\node[anchor=north, font=\scriptsize\bfseries] at ($(bratem.north)+(0,-.12)$) {Rate coding};
\begin{scope}[shift={(11.30,11.68)}]
    \draw[smallaxis] (0,0)--(1.68,0);
    \node[font=\tiny,anchor=north east] at (1.62,+.04) {drive};
    \draw[smallaxis] (0,0)--(0,.76);
    \node[font=\tiny,rotate=90] at (-.28,.39) {firing rate};
    \draw[posblue,line width=1.0pt] (.08,.06)--(.38,.06)..controls(.62,.09)and(1.06,.50)..(1.53,.66);
\end{scope}
\begin{scope}[shift={(11.30,8.70)}]
    \draw[smallaxis] (0,0)--(1.68,0);
    \node[font=\tiny,anchor=north east] at (1.62,+.04) {drive};
    \draw[smallaxis] (0,0)--(0,.76);
    \node[font=\tiny,rotate=90] at (-.28,.39) {firing rate};
    \draw[antorange,line width=1.0pt] (.08,.06)--(.38,.06)..controls(.62,.09)and(1.06,.50)..(1.53,.66);
\end{scope}

% Refractory renewal spikes with variable heights.
\node[blueblock, minimum width=3.25cm, minimum height=2.18cm] (bspikep) at (15.72,12.42) {};
\node[orangeblock, minimum width=3.25cm, minimum height=2.18cm] (bspikem) at (15.72,9.44) {};
\node[anchor=north, font=\scriptsize, align=center] at ($(bspikep.north)+(0,-.10)$)
    {\textbf{Refractory}\\[-0.4mm]\textbf{renewal process}};
\node[anchor=north, font=\scriptsize, align=center] at ($(bspikem.north)+(0,-.10)$)
    {\textbf{Refractory}\\[-0.4mm]\textbf{renewal process}};
\begin{scope}[shift={(14.55,11.58)}]
    \draw[smallaxis] (0,0)--(2.28,0);
    \foreach \x/\h in {.15/.24,.40/.54,.68/.34,.92/.70,1.22/.29,1.49/.48,1.79/.76,2.07/.39}
        {\draw[posblue,line width=.95pt] (\x,0)--(\x,\h);}
\end{scope}
\begin{scope}[shift={(14.55,8.60)}]
    \draw[smallaxis] (0,0)--(2.28,0);
    \foreach \x/\h in {.14/.42,.42/.25,.64/.68,.91/.36,1.18/.78,1.46/.50,1.77/.30,2.05/.61}
        {\draw[antorange,line width=.95pt] (\x,0)--(\x,\h);}
\end{scope}

\draw[bluearrow] (brecp.east)--(bratep.west);
\draw[bluearrow] (bratep.east)--(bspikep.west);
\draw[orangearrow] (brecm.east)--(bratem.west);
\draw[orangearrow] (bratem.east)--(bspikem.west);

% ============================================================
% (c) SPIKE TWITCH DECODING AND ACTUATOR MAPPING
% ============================================================
\node[panel, minimum width=18cm, minimum height=6.75cm, anchor=north west] (pc) at (0,7.08) {};
\node[anchor=west, font=\bfseries\large] at (0.15,6.73) {(c) Spike twitch decoding and actuator mapping};

% Large colored lanes.
\begin{scope}[on background layer]
    \fill[posblue!7, rounded corners=3pt] (0.45,3.98) rectangle (12.92,6.30);
    \fill[antorange!7, rounded corners=3pt] (0.45,0.86) rectangle (12.92,3.18);
\end{scope}

% Spike trains
\node[blueblock, minimum width=2.95cm, minimum height=2.02cm] (cspikep) at (1.88,5.14) {};
\node[orangeblock, minimum width=2.95cm, minimum height=2.02cm] (cspikem) at (1.88,2.02) {};
\node[anchor=north, font=\scriptsize\bfseries, text=posblue] at ($(cspikep.north)+(0,-.11)$) {Agonist spikes};
\node[anchor=north, font=\scriptsize\bfseries, text=antorange] at ($(cspikem.north)+(0,-.11)$) {Antagonist spikes};
\begin{scope}[shift={(0.78,4.42)}]
    \draw[smallaxis] (0,0)--(2.18,0);
    \foreach \x/\h in {.15/.26,.38/.60,.64/.36,.88/.76,1.13/.47,1.39/.31,1.68/.68,1.98/.42}
        {\draw[posblue,line width=.95pt] (\x,0)--(\x,\h);}
\end{scope}
\begin{scope}[shift={(0.78,1.30)}]
    \draw[smallaxis] (0,0)--(2.18,0);
    \foreach \x/\h in {.14/.49,.36/.28,.57/.72,.81/.38,1.06/.59,1.31/.24,1.60/.78,1.88/.44,2.05/.61}
        {\draw[antorange,line width=.95pt] (\x,0)--(\x,\h);}
\end{scope}

% Heterogeneous twitches -- larger boxes with labels separated from curves.
\node[blueblock, minimum width=5.00cm, minimum height=2.02cm] (ctop) at (6.15,5.14) {};
\node[orangeblock, minimum width=5.00cm, minimum height=2.02cm] (cbot) at (6.15,2.02) {};
\node[anchor=north, font=\scriptsize, align=center, text=posblue] at ($(ctop.north)+(0,-.10)$)
    {\textbf{Heterogeneous motor-unit}\\[-0.6mm]\textbf{twitches $\kappa$}\\[-0.5mm]{\tiny slow/weak to fast/strong}};
\node[anchor=north, font=\scriptsize, align=center, text=antorange] at ($(cbot.north)+(0,-.10)$)
    {\textbf{Heterogeneous motor-unit}\\[-0.6mm]\textbf{twitches $\kappa$}\\[-0.5mm]{\tiny slow/weak to fast/strong}};
\begin{scope}[shift={(4.05,4.40)}]
    \draw[smallaxis] (0,0)--(4.05,0);
    \draw[posblue!38,line width=.9pt] (.05,0)..controls(.45,.01)and(.82,.24)..(1.45,.30)..controls(2.10,.30)and(2.75,.05)..(3.35,.01);
    \draw[posblue!68,line width=.95pt] (.05,0)..controls(.28,.02)and(.48,.47)..(.96,.53)..controls(1.42,.53)and(1.72,.10)..(2.20,.02);
    \draw[posblue,line width=1.05pt] (.05,0)..controls(.16,.03)and(.25,.73)..(.55,.80)..controls(.84,.80)and(.98,.12)..(1.32,.02);
\end{scope}
\begin{scope}[shift={(4.05,1.28)}]
    \draw[smallaxis] (0,0)--(4.05,0);
    \draw[antorange!38,line width=.9pt] (.05,0)..controls(.45,.01)and(.82,.24)..(1.45,.30)..controls(2.10,.30)and(2.75,.05)..(3.35,.01);
    \draw[antorange!68,line width=.95pt] (.05,0)..controls(.28,.02)and(.48,.47)..(.96,.53)..controls(1.42,.53)and(1.72,.10)..(2.20,.02);
    \draw[antorange,line width=1.05pt] (.05,0)..controls(.16,.03)and(.25,.73)..(.55,.80)..controls(.84,.80)and(.98,.12)..(1.32,.02);
\end{scope}
\draw[bluearrow] (cspikep.east)--(ctop.west);
\draw[orangearrow] (cspikem.east)--(cbot.west);

% Summed pool outputs
\node[blueblock, minimum width=2.85cm, minimum height=2.02cm] (csump) at (10.82,5.14) {};
\node[orangeblock, minimum width=2.85cm, minimum height=2.02cm] (csumm) at (10.82,2.02) {};
\node[anchor=north, font=\scriptsize, align=center, text=posblue] at ($(csump.north)+(0,-.11)$)
    {\textbf{Summed pool}\\[-0.4mm]\textbf{output $z^{+}_{t,m}$}};
\node[anchor=north, font=\scriptsize, align=center, text=antorange] at ($(csumm.north)+(0,-.11)$)
    {\textbf{Summed pool}\\[-0.4mm]\textbf{output $z^{-}_{t,m}$}};
\begin{scope}[shift={(9.80,4.40)}]
    \draw[smallaxis] (0,0)--(2.02,0);
    \draw[posblue,line width=1.0pt] (.03,.02)..controls(.13,.02)and(.21,.55)..(.42,.55)
      ..controls(.61,.55)and(.70,.22)..(.86,.27)..controls(1.02,.33)and(1.13,.09)..(1.29,.08)
      ..controls(1.50,.06)and(1.64,.22)..(1.82,.20)..controls(1.94,.17)and(1.98,.05)..(2.00,.02);
\end{scope}
\begin{scope}[shift={(9.80,1.28)}]
    \draw[smallaxis] (0,0)--(2.02,0);
    \draw[antorange,line width=1.0pt] (.03,.02)..controls(.17,.02)and(.29,.42)..(.51,.44)
      ..controls(.70,.44)and(.80,.20)..(.96,.15)..controls(1.11,.10)and(1.25,.11)..(1.40,.14)
      ..controls(1.61,.17)and(1.72,.31)..(1.88,.27)..controls(1.96,.22)and(1.99,.07)..(2.00,.02);
\end{scope}
\draw[bluearrow] (ctop.east)--(csump.west);
\draw[orangearrow] (cbot.east)--(csumm.west);

% Actuator mapping
\node[block, minimum width=4.15cm, minimum height=5.0cm] (cmap) at (15.55,3.58) {};
\node[anchor=north, font=\small\bfseries] at ($(cmap.north)+(0,-.05)$) {Actuator mapping};
\node[block, minimum width=3.45cm, minimum height=1.8cm] (ctorque) at (15.55,4.5) {
    {\scriptsize\textbf{Torque Ant}}\\[-0.2mm]
    {\scriptsize (signed perturbation)}\\[+1.2mm]
    $z_{t,m}=z^{+}_{t,m}-z^{-}_{t,m}$
};
\node[block, minimum width=3.45cm, minimum height=1.8cm] (cmuscle) at (15.55,2.5) {
    {\scriptsize\textbf{Agonist/Antagonist Ant}}\\[-0.2mm]
    {\scriptsize (separate channels)}\\[+1.2mm]
    $\left[\begin{array}{c} z^{+}_{t,m}\\ z^{-}_{t,m}\end{array}\right]$
};

% Merge the two pool outputs visually before the common mapping block.
\coordinate (merge) at (13.45,3.58);
\draw[bluearrow] (csump.east) -- ++(.50,0) |- (merge);
\draw[orangearrow] (csumm.east) -- ++(.50,0) |- (merge);

\end{tikzpicture}%
}
\caption{Overview of the Poisson and Spike proposal distributions.
(a) Poisson generates sparse signed events from fixed positive and negative event rates and recruitment marks, then decodes the events with a common twitch kernel.
(b) Spike uses a shared stochastic drive for agonist and antagonist motor pools. Increasing drive recruits motor units progressively in ordered steps, while derecruitment occurs at lower drive levels, producing hysteresis. Active units then use rate coding and refractory renewal discharge; spike heights illustrate different motor-unit strengths.
(c) Motor-unit-specific twitch responses are summed separately for the agonist and antagonist pools. The resulting outputs are converted either to a signed torque perturbation or to separate agonist and antagonist actuator channels.}
\label{fig:poisson-spike-overview}
\end{figure*}

Beyond continuous Gaussian perturbations, information-theoretic MPC has been extended to systems with compound-Poisson jump processes \cite{wang2019jump}, demonstrating that discontinuous event statistics can be incorporated into path-integral control. Such processes provide an alternative representation of stochastic exploration: event times determine \emph{when} a perturbation occurs, while associated marks determine its direction and magnitude. Separating event occurrence from event effect offers a natural means of constructing sparse, temporally structured control proposals. In particular, event-based representations allow temporal organization to emerge from the interaction between discrete events and their continuous-time responses rather than solely from the covariance of a Gaussian process.

The spinal motoneuron pools that drive skeletal muscle provide a natural source of such structure. Motoneuron populations transform shared neural drive into coordinated muscle activation through ordered recruitment, motor-unit discharge, and temporally extended twitch responses \cite{henneman1957relation, hug2023common, fuglevand1993models, caillet2023motoneuron}. Recruitment thresholds, firing-rate modulation, and heterogeneous motor-unit responses jointly determine how common neural inputs are converted into mechanical output. These mechanisms suggest a way to construct non-Gaussian control proposals with structured temporal dependencies, without prescribing a gait, reference trajectory, or learned control policy.

Motivated by these observations, we introduce two event-driven proposal distributions for MPPI (Fig.~\ref{fig:poisson-spike-overview}). The first is a sparse marked Poisson proposal in which signed recruitment events are decoded through a causal twitch kernel (Fig.~\ref{fig:noise-exploration}, orange). The second, Spike-MPPI, uses a motoneuron-inspired motor-pool model with common stochastic drive, ordered recruitment, recruitment--derecruitment hysteresis, rate coding, refractory renewal discharge, and heterogeneous motor-unit twitch dynamics (Fig.~\ref{fig:noise-exploration}, green). Both proposals generate temporally coherent perturbations, but differ in how events are produced and organized. The Poisson model provides a simpler event-based reference, whereas Spike incorporates the recruitment and discharge mechanisms of interacting motor-unit populations.

Our experimental comparison is designed to examine the contribution of the proposal distribution independently of changes to the broader controller. All sampling methods use the same predictive dynamics, task objectives, control constraints, temperature adaptation, and MPPI update rule, with matched nominal exploration magnitudes and rollout budgets. We compare a temporally correlated Gaussian reference (Standard), pink noise with long-range temporal correlations, the marked Poisson proposal, and Spike-MPPI. In addition, Gaussian-Poisson and Gaussian-Spike controls reproduce the estimated finite-horizon spectra of their corresponding event-driven proposals, allowing us to examine whether matching second-order spectral properties is sufficient to recover their closed-loop behavior. These comparisons probe the effects of temporal spectral shaping and additional features of event-driven sampling, although spectral matching does not reproduce all nonstationary dependencies or isolate individual motor-pool mechanisms. The aim is therefore to characterize the effects of sampling structure rather than to establish superiority over alternative MPPI formulations.

We evaluate these sampling methods on a planar ablation of the motor-pool mechanisms and on locomotion and manipulation tasks using two MuJoCo \cite{todorov2012mujoco} Ant actuation models and a Franka Research 3 (FR3) manipulator, with additional reaching experiments on a physical FR3. The locomotion models comprise the original eight-channel torque-driven Ant and a sixteen-channel agonist--antagonist variant in which simultaneous activation modulates joint impedance. The manipulation experiments include reaching, box pushing, ball balancing, and ball bouncing. Across these systems, we examine task performance, control smoothness, and computational cost, together with inter-joint coordination and, for antagonistic actuation, co-contraction and differential activation.

\section{Methodology}
\label{sec:methodology}

\subsection{Problem Formulation}
\label{sec:problem-formulation}

We consider online receding-horizon control of an articulated dynamical system with state $\mathbf{x}_t\in\mathcal{X}\subset\mathbb{R}^{n_x}$ and actuator command $\mathbf{u}_t\in\mathcal{U}\subset\mathbb{R}^{n_u}$. The discrete-time predictive dynamics are
\begin{equation}
    \mathbf{x}_{t+1}=f(\mathbf{x}_t,\mathbf{u}_t),
    \label{eq:system-dynamics}
\end{equation}
where $f$ represents the predictive dynamics, including nonlinear rigid-body motion, contacts, actuator limits, and task-dependent interactions. At control update $k$, the controller optimizes a sequence
\begin{equation}
    \mathbf{U}=[\mathbf{u}_0,\ldots,\mathbf{u}_{H-1}]
    \label{eq:control-sequence}
\end{equation}
over a horizon of $H$ control steps to minimize
\begin{equation}
    J(\mathbf{U};\mathbf{x}_k)=
    \sum_{t=0}^{H-1}\ell(\mathbf{x}_t,\mathbf{u}_t)+J_f(\mathbf{x}_H),
    \label{eq:control-objective}
\end{equation}
where $\ell$ and $J_f$ denote running and terminal costs. Model Predictive Path Integral (MPPI) control samples candidate sequences around a nominal plan $\overline{\mathbf{U}}_k$. Its initial plan repeats a robot-specific neutral command:
\begin{equation}
    \overline{\mathbf{U}}_0=
    [\mathbf{u}_{\mathrm{neutral}},\ldots,\mathbf{u}_{\mathrm{neutral}}].
    \label{eq:initial-nominal}
\end{equation}
The neutral command is zero for torque-controlled locomotion; for the position-controlled Franka Research 3 (FR3), the neutral command is initialized from the current commanded joint configuration. Subsequent updates are warm-started by shifting the previous solution and repeating its last command,
\begin{equation}
    \overline{\mathbf{U}}_{k+1}
    =[\mathbf{u}^{\star}_{k,1},\ldots,
      \mathbf{u}^{\star}_{k,H-1},\mathbf{u}^{\star}_{k,H-1}].
    \label{eq:warm-start}
\end{equation}
Thus, after initialization, each sampler explores around the controller's previously optimized sequence.

\subsection{Model Predictive Path Integral Control}
\label{sec:mppi}

At control update $k$, MPPI draws $N$ perturbation sequences $\boldsymbol{\epsilon}_k^{(i)}$ and constructs candidate actuator commands
\begin{equation}
    \mathbf{U}_k^{(i)}=
    \overline{\mathbf{U}}_k+\boldsymbol{\epsilon}_k^{(i)},\qquad i=1,\ldots,N.
    \label{eq:mppi-candidates}
\end{equation}
Each candidate is then propagated through the predictive dynamics and assigned a trajectory cost $J_i$, with $J_i=\infty$ for invalid rollouts. Let $\mathcal V=\{i:J_i<\infty\}$ be the set of valid rollouts and $\rho_J=\min_{i\in\mathcal V}J_i$. The normalized exponential weights are
\begin{equation}
    w_i=
    \frac{\exp\!\left[-(J_i-\rho_J)/\lambda\right]}
         {\sum_{j\in\mathcal V}\exp\!\left[-(J_j-\rho_J)/\lambda\right]},
    \qquad i\in\mathcal V,
    \label{eq:mppi-weights}
\end{equation}
and $w_i=0$ for $i\notin\mathcal V$, where $\lambda>0$ is the sampling temperature. If $\mathcal V$ is empty, the applied command is the previous plan. The updated plan is the cost-weighted average of the admissible candidate commands,
\begin{equation}
    \mathbf{U}_k^{\star}=\sum_{i=1}^{N}w_i\mathbf{U}_k^{(i)}.
    \label{eq:mppi-update}
\end{equation}
The first command of $\mathbf{U}_k^{\star}$ is applied to the plant; the horizon is then shifted, and the procedure is repeated. The temperature is adapted online from the valid costs using Lower-Bound Policy Search (LBPS) \cite{watson2023inferring}. All sampling methods share the same dynamics, task objective, control limits, temperature adaptation, and weighted-update rule within each experimental condition. They differ only in the distribution of the sampled perturbations. The common exponential weighting provides a controlled comparison of proposal distributions; it is not claimed to be an importance-corrected estimator for an unchanged path-integral reference measure.

\subsection{Gaussian Sampling}
\label{sec:gaussian-proposal}

Independent Gaussian sampling provides the reference exploration distribution underlying the Gaussian MPPI framework \cite{williams2015model}. In proposal coordinates, its perturbations satisfy
\begin{equation}
    \boldsymbol{\epsilon}_{i,t}\sim
    \mathcal{N}(\mathbf{0},\boldsymbol{\Sigma}),
    \label{eq:gaussian-white}
\end{equation}
where $\boldsymbol{\Sigma}$ is diagonal and determines the coordinate-specific exploration variance. Samples are independent across timesteps and rollouts.

\subsection{Temporally Correlated Gaussian Sampling}
\label{sec:standard-proposal}

Temporal continuity is introduced using a stationary first-order autoregressive (AR(1)) filter, a simple temporally correlated Gaussian proposal related to those used in sampling-based trajectory optimization \cite{pinneri2021sample}:
\begin{equation}
    \boldsymbol{\epsilon}_{i,t}
    =\rho_s\boldsymbol{\epsilon}_{i,t-1}
    +\sqrt{1-\rho_s^2}\,\boldsymbol{\xi}_{i,t},
    \qquad \boldsymbol{\xi}_{i,t}\sim
    \mathcal{N}(\mathbf{0},\boldsymbol{\Sigma}),
    \label{eq:standard-ar1}
\end{equation}
with the initial perturbation drawn from $\mathcal{N}(\mathbf{0},\boldsymbol{\Sigma})$. Consequently,
\begin{equation}
    \operatorname{Cov}(\boldsymbol{\epsilon}_{i,t},
      \boldsymbol{\epsilon}_{i,t+\tau})
      =\rho_s^{|\tau|}\boldsymbol{\Sigma}.
    \label{eq:standard-covariance}
\end{equation}
We use $\rho_s=0.25$. This proposal is referred to hereafter as \emph{Standard}. It serves as the implemented Gaussian MPPI baseline, following the sampling-based framework of Williams et al.~\cite{williams2015model}.

\subsection{Pink Noise Sampling}
\label{sec:pink-proposal}

Pink noise provides a Gaussian proposal with increased power at low frequencies. Its idealized power spectral density obeys $S_{\mathrm{pink}}(f)\propto 1/f$ for $f>0$, a form of colored-noise exploration studied in continuous control \cite{eberhard2023pink}. Unlike the exponential correlation of Standard, this spectrum emphasizes variations that persist across the planning horizon.

For each rollout and proposal coordinate, independent white-Gaussian samples $\boldsymbol{\xi}_{i,t}\sim\mathcal{N}(\mathbf{0},\mathbf{I})$ are transformed along the horizon,
\begin{equation}
    \mathbf{X}_{i,k}=\mathcal{F}\{\boldsymbol{\xi}_{i,t}\}_k.
    \label{eq:pink-fft}
\end{equation}
Writing $f_k=k/H$ for the nonnegative discrete frequencies, the amplitude filter is
\begin{equation}
    G_k=
    \begin{cases}
        f_1^{-1/2}, & k=0,\\
        f_k^{-1/2}, & k\geq 1.
    \end{cases}
    \label{eq:pink-filter}
\end{equation}
The finite DC amplitude avoids forcing each sampled trajectory to have zero mean, allowing sustained positive or negative perturbations. The filter is normalized by the expected time-domain variance, and the resulting samples are
\begin{equation}
    \boldsymbol{\epsilon}_{i,t}=
    \boldsymbol{\Sigma}^{1/2}\,
    \mathcal{F}^{-1}\{\widetilde G_k\mathbf X_{i,k}\}_t,
    \label{eq:pink-sampling}
\end{equation}
where $\widetilde G_k$ denotes the normalized filter. Each rollout and proposal coordinate is sampled independently before temporal filtering. This FFT construction yields a finite-horizon Gaussian process with $1/f$ power at nonzero discrete frequencies and a regularized DC component.

\subsection{Poisson Sampling with Twitch Kernels}
\label{sec:poisson-proposal}

The Poisson proposal replaces Gaussian perturbations with marked point-process events followed by a causal twitch decoder. Each signed proposal coordinate $j$ has independent positive and negative event streams. For rollout $i$, prediction step $t$, and recruitment-depth mark $r\in\{1,\ldots,L\}$,
\begin{align}
    N^{+}_{i,t,j,r}&\sim\operatorname{Poisson}\!\left(
       \frac{\lambda_0}{2}\Delta t\,p_r\right),\\
    N^{-}_{i,t,j,r}&\sim\operatorname{Poisson}\!\left(
       \frac{\lambda_0}{2}\Delta t\,p_r\right),
    \label{eq:poisson-events}
\end{align}
where $p_r=1/L$. The recruitment mark encodes cumulative activation of an ordered motor-unit pool. Unit $\ell$ has normalized strength
\begin{equation}
    g_{\ell}=\frac{2\ell}{L(L+1)},\qquad \ell=1,\ldots,L,
    \label{eq:unit-strength}
\end{equation}
so that $\sum_{\ell=1}^{L}g_{\ell}=1$. An event of depth $r$ recruits units $1,\ldots,r$ and has amplitude
\begin{equation}
    a_r=\sum_{\ell=1}^{r}g_{\ell}
       =\frac{r(r+1)}{L(L+1)}.
    \label{eq:cumulative-recruitment-mark}
\end{equation}
The positive and negative impulse trains are
\begin{equation}
    y^{p}_{i,t,j}=\sum_{r=1}^{L}a_rN^{p}_{i,t,j,r},
    \qquad p\in\{+,-\}.
    \label{eq:poisson-impulse-trains}
\end{equation}
Each event is decoded through a causal bi-exponential twitch kernel. For $q=0,\ldots,Q-1$, its unnormalized discrete samples are $\widetilde{\kappa}_q=\exp[-(q+1)\Delta t/\tau_d]-\exp[-(q+1)\Delta t/\tau_r]$, where $\tau_d>\tau_r$ are the decay and rise time constants. The peak-normalized kernel is $\kappa_q=\widetilde{\kappa}_q/\max_{0\le v<Q}\widetilde{\kappa}_v$. The decoded pool outputs are
\begin{equation}
    z^{p}_{i,t,j}=\sum_{q=0}^{Q-1}\kappa_qy^{p}_{i,t-q,j},
    \label{eq:poisson-decoding}
\end{equation}
with events before the start of the horizon set to zero. For signed actuator or Cartesian proposal coordinates, the raw perturbation is $z^+-z^-$. For the antagonistic muscle model, $z^+$ and $z^-$ are instead routed to their corresponding nonnegative actuator channels, and the rollout-population mean is removed at each horizon coordinate:
\begin{equation}
    \widetilde{\mathbf z}^{(i)}_t
       =\mathbf z^{(i)}_t-
        \frac{1}{N}\sum_{n=1}^{N}\mathbf z^{(n)}_t.
    \label{eq:population-centering}
\end{equation}
The symmetric signed construction has zero mean in expectation and requires no analogous population centering. Before actuator clipping, the Poisson proposal is scaled to match the nominal root-mean-square exploration magnitude of Standard. Because clipping changes the distribution near control limits, post-clipping variances may not be identical.

\subsection{Spike Sampling}
\label{sec:spike-proposal}

\begin{figure}[!t]
    \centering
    \includegraphics[width=\linewidth]{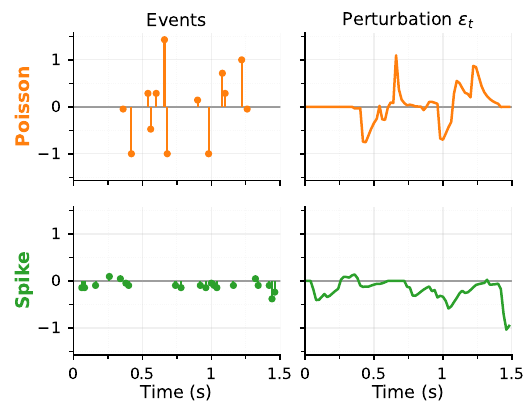}
    \caption{Representative Poisson (top) and Spike (bottom) perturbations. Left: Poisson events and Spike motor-unit discharges. Right: resulting perturbations after the common twitch kernel (Poisson) and the heterogeneous twitch kernels (Spike).}
    \label{fig:sampling-noise-example}
\end{figure}

Spike sampling replaces the fixed-rate Poisson law with a reduced-order stochastic motor-pool model (Fig.~\ref{fig:sampling-noise-example}). It is not intended to reproduce complete spinal electrophysiology, but to embed common drive, ordered recruitment, hysteresis, rate coding, refractory renewal discharge, and heterogeneous twitches in the sampling distribution \cite{henneman1957relation,fuglevand1993models,hug2023common,caillet2023motoneuron}, thereby introducing richer physiological and temporal structure than the simpler Poisson proposal.

\paragraph{Antagonistic pools and common drive}
Each signed proposal coordinate $j$ has agonist $(+)$ and antagonist $(-)$ pools. A first-order stochastic process generates a common drive across the units of each pair,
\begin{equation}
    s_{i,t+1,j}=\rho_d s_{i,t,j}
       +\sigma_d\sqrt{1-\rho_d^2}\,\xi_{i,t,j},
    \label{eq:spike-common-drive}
\end{equation}
with $\xi_{i,t,j}\sim\mathcal{N}(0,1)$ and $\rho_d=\exp(-\Delta t/\tau_d^{\mathrm{drive}})$. Half-wave rectification gives
\begin{align}
    d^{+}_{i,t,j}&=d_{\mathrm{co}}+[s_{i,t,j}]_+,\\
    d^{-}_{i,t,j}&=d_{\mathrm{co}}-[s_{i,t,j}]_-,
    \label{eq:spike-antagonistic-drive}
\end{align}
where $[x]_+=\max(x,0)$ and $[x]_- =\min(x,0)$, and both drives are capped at $1$. The torque-driven robot uses $d_{\mathrm{co}}=0$; the antagonistic actuator model uses $d_{\mathrm{co}}=0.08$, permitting concurrent pool activity and physical co-contraction. The model reflects qualitative features of common neural drive and antagonist co-activation rather than a subject-specific physiological fit \cite{farina2014effective,hug2023common,lewis2010cocontraction}.

\paragraph{Ordered recruitment and hysteresis}
Each pool contains $L$ units with normalized index $\phi_\ell=(\ell-1)/(L-1)$. Their recruitment thresholds follow
\begin{equation}
    \theta^{\mathrm{on}}_{\ell}=\theta_{\min}
       \left(\frac{\theta_{\max}}{\theta_{\min}}\right)^{\phi_\ell},
    \label{eq:spike-recruitment-thresholds}
\end{equation}
where $\theta_{\min}=0.10$ and $\theta_{\max}=0.85$. Derecruitment occurs at a lower threshold,
\begin{equation}
    \theta^{\mathrm{off}}_{\ell}=
       \theta^{\mathrm{on}}_{\ell}-0.06-0.08\theta^{\mathrm{on}}_{\ell}.
    \label{eq:spike-derecruitment-thresholds}
\end{equation}
Let $b^{p}_{i,t,j,\ell}\in\{0,1\}$ be the recruitment state. Its update is
\begin{equation}
    b^{p}_{i,t,j,\ell}=
    \begin{cases}
       1,&d^p_{i,t,j}\geq\theta^{\mathrm{on}}_{\ell},\\
       0,&d^p_{i,t,j}\leq\theta^{\mathrm{off}}_{\ell},\\
       b^{p}_{i,t-1,j,\ell},&\text{otherwise}.
    \end{cases}
    \label{eq:spike-hysteresis-state}
\end{equation}
Thus, a recruited unit can remain active after the drive falls below its recruitment threshold \cite{heckman2008persistent}. Unit strengths follow Eq.~\eqref{eq:unit-strength}. The background drive is below the lowest activation threshold yet above the lowest derecruitment threshold, allowing previously active antagonist units to overlap through hysteresis.

\paragraph{Rate coding and renewal discharge}
For a recruited unit, define the normalized suprathreshold drive
\begin{equation}
    r^{p}_{i,t,j,\ell}=\operatorname{clip}\!\left(
       \frac{d^p_{i,t,j}-\theta^{\mathrm{on}}_{\ell}}
            {1-\theta^{\mathrm{on}}_{\ell}},0,1\right).
    \label{eq:spike-rate-drive}
\end{equation}
The firing rate is
\begin{equation}
    f^{p}_{i,t,j,\ell}=b^{p}_{i,t,j,\ell}
       \left[f_{\min}+(f_{\max}-f_{\min})r^{p}_{i,t,j,\ell}\right],
    \label{eq:spike-firing-rate}
\end{equation}
where $f_{\min}=\lambda_0$ and $f_{\max}=3\lambda_0$. Rather than memoryless Poisson discharge, active units generate spikes with refractory shifted-Gamma inter-spike intervals \cite{fuglevand1993models,navallas2019gamma},
\begin{equation}
    \Delta T_{\ell}=\tau_{\mathrm{ref}}+G_{\ell},\qquad
    G_{\ell}\sim\Gamma\!\left(
      \alpha_{\Gamma},\frac{1/f_{\ell}-\tau_{\mathrm{ref}}}{\alpha_{\Gamma}}\right),
    \label{eq:spike-renewal-discharge}
\end{equation}
where the second Gamma parameter is the scale, $\alpha_{\Gamma}=3$ is the shape, and $\tau_{\mathrm{ref}}=5\,\mathrm{ms}$. An initially randomized discharge countdown avoids artificial synchronization at recruitment. For time-varying drive, $f_{\ell}$ in Eq.~\eqref{eq:spike-renewal-discharge} denotes the instantaneous firing rate used when the next interval is drawn. At every control update, each rollout and proposal coordinate starts with all units inactive and a drive drawn from the stationary distribution of Eq.~\eqref{eq:spike-common-drive}; no sampler state is carried between updates. These parameters model qualitative refractory firing rather than measured subject-specific neural physiology \cite{borg1983prior}.

\paragraph{Heterogeneous motor-unit twitches}
Later-recruited units have larger strengths $g_{\ell}$ and shorter twitch times. Their rise and decay constants are
\begin{align}
    \tau_{r,\ell}&=\tau_r^0(1.30-0.55\phi_{\ell}),\\
    \tau_{d,\ell}&=\tau_d^0(1.45-0.65\phi_{\ell}).
    \label{eq:spike-heterogeneous-times}
\end{align}
If $\mathcal S^{p}_{i,j,\ell}$ denotes the discharge bins of unit $\ell$ in pool $p$, the pool output is
\begin{equation}
    z^{p}_{i,t,j}=\sum_{\ell=1}^{L}g_{\ell}
       \sum_{k\in\mathcal S^p_{i,j,\ell}}\kappa_{\ell,t-k},
    \qquad p\in\{+,-\},
    \label{eq:spike-pool-output}
\end{equation}
where $\kappa_{\ell}$ denotes the unit-specific twitch kernel obtained using the rise and decay constants in Eq.~\eqref{eq:spike-heterogeneous-times}. Kernel values with negative lag are zero. This creates temporally extended, motor-unit-dependent perturbations \cite{fuglevand1993models,caillet2023motoneuron}. Signed proposal coordinates use $z^+-z^-$; the antagonistic Ant retains the two pools as separate channels and centers them according to Eq.~\eqref{eq:population-centering}. Like Poisson, Spike is scaled to match Standard's pre-clipping exploration RMS without removing its distinct higher-order statistics.

\subsection{Spectrum-Matched Gaussian Sampling}
\label{sec:spectrum-matched-gaussian}

To examine which effects can be explained by temporal frequency content, we use two spectrum-matched Gaussian controls: Gaussian-Poisson and Gaussian-Spike. Each reproduces the estimated finite-horizon second-order spectrum of its corresponding structured sampler, while not reproducing its event sparsity, firing history, or non-Gaussian amplitude statistics.

For a target proposal $q\in\{\mathrm{Poisson},\mathrm{Spike}\}$, we draw $M_{\mathrm{cal}}$ independent reference perturbations using the same horizon and pre-clipping scaling. Let $\mathbf z^{(r)}_{t,m}\in\mathbb R^{d_m}$ denote realization $r$ for proposal group $m$, with $d_m=1$ for a signed actuator or Cartesian coordinate and $d_m=2$ for an antagonistic pair. Its ensemble-centered value and Fourier transform are
\begin{align}
    \widetilde{\mathbf z}^{(r)}_{t,m}
      &=\mathbf z^{(r)}_{t,m}
       -\frac{1}{M_{\mathrm{cal}}}\sum_{s=1}^{M_{\mathrm{cal}}}
             \mathbf z^{(s)}_{t,m},\\
    \mathbf Z^{(r)}_{k,m}
      &=\sum_{t=0}^{H-1}\widetilde{\mathbf z}^{(r)}_{t,m}
             \exp\!\left(-\mathrm{i}\frac{2\pi kt}{H}\right).
    \label{eq:spectrum-calibration}
\end{align}
The estimated cross-spectral density matrix is
\begin{equation}
    \widehat{\mathbf S}_{k,m}=
      \frac{1}{M_{\mathrm{cal}}}\sum_{r=1}^{M_{\mathrm{cal}}}
        \mathbf Z^{(r)}_{k,m}\mathbf Z^{(r)H}_{k,m}.
    \label{eq:spectrum-target}
\end{equation}
For a signed channel this is a scalar power spectrum; for an antagonistic pair it is a $2\times2$ matrix that also captures second-order coupling between the two channels. We obtain a spectral factor from the Hermitian eigendecomposition such that
\begin{equation}
    \mathbf A_{k,m}\mathbf A_{k,m}^{H}=
         \frac{1}{H}\widehat{\mathbf S}_{k,m}.
    \label{eq:spectrum-factor}
\end{equation}
At runtime, white-Gaussian sequences $\mathbf w_{t,m}\sim\mathcal N(\mathbf 0,\mathbf I)$ are Fourier transformed to $\mathbf W_{k,m}=\mathcal F\{\mathbf w_{t,m}\}_k$, filtered, and inverted:
\begin{equation}
    \mathbf Y_{k,m}=\mathbf A_{k,m}\mathbf W_{k,m},\qquad
    \boldsymbol\epsilon_{t,m}=\mathcal F^{-1}\{\mathbf Y_{k,m}\}_t.
    \label{eq:spectrum-matched-sampling}
\end{equation}
Conjugate symmetry $\mathbf Y_{H-k,m}=\mathbf Y_{k,m}^{*}$ is enforced at paired frequencies, with real DC and (for even $H$) Nyquist bins. Under the unnormalized DFT convention, $\mathbb E[\mathbf W_{k,m}\mathbf W_{k,m}^{H}]=H\mathbf I$, giving
\begin{equation}
    \mathbb E[\mathbf Y_{k,m}\mathbf Y_{k,m}^{H}]
       =\widehat{\mathbf S}_{k,m}.
    \label{eq:spectrum-matching-property}
\end{equation}
Thus, the generated Gaussian perturbations reproduce the estimated target spectrum in expectation before control clipping. Finite-horizon Fourier matching does not necessarily reproduce all nonstationary covariance structure or higher-order dependencies of the original event process.

\subsection{Robot Models}
\label{sec:robot-models}

\subsubsection{Locomotion}
We evaluate the samplers on two actuation models sharing the same Ant rigid-body morphology, joint topology, environment, and task objective. They differ in joint actuation and consequently in their actuation-dependent dynamics.

\paragraph{Torque-driven Ant}
The standard MuJoCo/Gymnasium Ant has eight bidirectional motors and control vector
\begin{equation}
    \mathbf u\in[-1,1]^8.
\end{equation}
The actuator transmission gain gives a nominal peak generalized-force magnitude $F_0=150$ at $|u_j|=1$. Standard and Pink perturb eight actuator coordinates; Poisson generates eight signed twitch-decoded channels; and Spike takes the difference between the agonist- and antagonist-pool outputs to generate eight-dimensional commands.

\paragraph{Agonist/antagonist Ant}
The second model preserves the same rigid-body system but replaces each bidirectional motor with an agonist/antagonist pair, giving 16 nonnegative commands
\begin{equation}
    a_j^+,a_j^-\in[0,1],\qquad j=1,\ldots,8.
    \label{eq:antagonist-controls}
\end{equation}
The MuJoCo actuators use an instantaneous command-to-force mapping, without an additional plant activation filter. Standard and Pink sample the 16 actuator coordinates; Poisson uses a common twitch kernel; and Spike uses heterogeneous motor-unit twitches. For each joint,
\begin{equation}
    \tau_j=\tau_0(a_j^+-a_j^-)
       -K(a_j^++a_j^-)q_j
       -B(a_j^++a_j^-)\dot q_j,
    \label{eq:ant-bio-torque}
\end{equation}
where $\tau_0=150\,\mathrm{N\,m}$, $K=18\,\mathrm{N\,m/rad}$, and $B=1.5\,\mathrm{N\,m\,s/rad}$. Differential activation produces directional torque, while simultaneous activation modulates stiffness and damping. Equal activation cancels the directional term but retains the impedance terms, so co-contraction has a mechanical effect. Poisson routes positive and negative events to their corresponding actuator channels; Spike directly produces two motor-pool outputs.

\subsubsection{Manipulation}
\paragraph{Franka Research 3}
Manipulation experiments use a seven-degree-of-freedom Franka Research 3 (FR3) arm with joint-position-controlled actuators, simulated with the MuJoCo Menagerie Franka Emika Panda model and its joint and velocity limits. Unlike locomotion, sampling perturbations are defined in three-dimensional Cartesian end-effector position space, not directly in joint space. They are mapped to joint-position offsets by differential inverse kinematics,
\begin{equation}
    \Delta\mathbf q_{i,t}=
       \mathbf J(\overline{\mathbf q}_t)^{\dagger}
       \Delta\mathbf x_{i,t},
    \label{eq:fr3-ik}
\end{equation}
where $\Delta\mathbf x_{i,t}\in\mathbb R^3$ is the sampled Cartesian perturbation and $\mathbf J\in\mathbb R^{3\times7}$ is the translational end-effector Jacobian evaluated along the nominal joint-position sequence. The Moore--Penrose pseudoinverse is computed numerically with relative singular-value cutoff $10^{-4}$. The resulting offsets are added to the nominal joint positions and clipped to the actuator bounds. Rollouts, costs, and the MPPI weighted update operate in joint space.

\subsection{Tasks}
\label{sec:tasks}

The rollout objectives primarily emphasize task completion rather than explicitly optimizing control smoothness, energy consumption, coordination, or co-contraction. Small task-dependent stabilization and regularization terms are nevertheless retained, and their weights are held fixed across sampling methods within a given experimental condition. Smoothness and coordination are therefore evaluated as outcomes rather than directly targeted optimization criteria.

\subsubsection{Locomotion}
All locomotion experiments use the same closed stadium track. Let $s(\mathbf p)$ be the periodic track-progress coordinate of planar point $\mathbf p$, with $\Delta s_t$ the signed wrapped progress increment between consecutive controller steps. Depending on the task, the primary body is the robot torso, a pushed box, or a towed sled. A lap is completed once the cumulative progress of this body increases by one track length.

\begin{figure*}[!t]
    \centering
    \includegraphics[width=\textwidth]{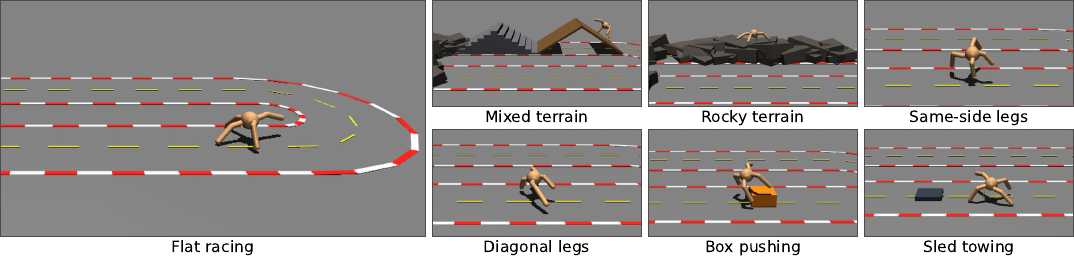}
    \caption{Locomotion scenarios. Left: flat racing track. Right: generalization scenarios with mixed and rocky terrain, same-side and diagonal leg-length asymmetries, box pushing, and sled towing.}
    \label{fig:locomotion-scenes}
\end{figure*}

\paragraph{Running}
In the baseline locomotion task, the torso is the primary tracked body. The controller is rewarded for forward progress while remaining upright and on the road. Flat running (Fig.~\ref{fig:locomotion-scenes}, left) is used for the main rollout-budget comparison and also serves as the baseline for terrain, morphology, and planner-mismatch studies.

\paragraph{Box pushing}
A free box is initially placed $1.8\,\mathrm{m}$ ahead of the robot. Its footprint is $0.90\times0.90\,\mathrm{m}$, its height $0.45\,\mathrm{m}$, its mass $0.1\,\mathrm{kg}$, and its friction parameter $0.60$. Box progress determines lap completion. Dense shaping additionally rewards coupled robot--box progress and decreasing robot--box distance before contact. A candidate is rejected if the box rises more than $0.12\,\mathrm{m}$ above its resting height or its up-axis component drops below $0.75$.

\paragraph{Sled towing}
A sled with dimensions $1.0\times0.80\times0.16\,\mathrm{m}$, mass $0.1\,\mathrm{kg}$, and friction parameter $0.60$ is placed $1.8\,\mathrm{m}$ ahead and connected to the robot by a $3.0\,\mathrm{m}$ rope. Sled progress determines lap completion. Because the rope mechanically couples the bodies, no approach-to-object shaping term is used. The same $0.12\,\mathrm{m}$ lift limit applies, with a minimum up-axis threshold of $0.70$.

\paragraph{Rollout objective}
The objective rewards task progress, with small stabilization and control-deviation terms. Let $P_t$ be cumulative progress of the primary body and $R_t$ that of the robot root. For the object tasks, define $C_t=\min\!\{R_t,\max(P_t,0)\}$ and $A_t=D_0-D_t$, with $D_t$ the planar robot--object distance. Let $h_t$ be the torso up-axis component, $\overline{\mathbf u}_t$ the nominal command, and $\mathbf s_u$ the actuator control-scale vector. The finite-horizon cost is
\begin{align}
    J={}&-\frac{w_p}{H}\sum_{t=1}^{H}P_t
          -\frac{w_r}{H}\sum_{t=1}^{H}C_t
          -\frac{w_a}{H}\sum_{t=1}^{H}A_t \nonumber\\
       &+w_{\mathrm{up}}\sum_{t=1}^{H}(1-h_t)^2
          +\frac{w_u}{n_u}\sum_{t=1}^{H}
          \left\| (\mathbf u_t-\overline{\mathbf u}_t)
                     \oslash\mathbf s_u\right\|_2^2,
    \label{eq:task-rollout-cost}
\end{align}
where $\oslash$ denotes elementwise division. We set $w_{\mathrm{up}}=0.05$ and $w_u=10^{-4}$ for all three tasks. For running, the primary body is the robot itself, so only the primary progress term is active; the task-dependent weights are reported in Table~\ref{tab:task-weights}. Invalid rollouts receive infinite cost when the task body or root leaves the road corridor or when the robot or manipulated object falls.

\begin{table}[!htbp]
    \centering
    \caption{Task-specific locomotion rollout-cost weights.}
    \label{tab:task-weights}
    \begin{tabular}{lccc}
        \toprule
        Task & $w_p$ & $w_r$ & $w_a$ \\
        \midrule
        Run      & 1.00 & 0    & 0    \\
        Push box & 1.00 & 0.35 & 1.00 \\
        Tow sled & 1.00 & 0.25 & 0    \\
        \bottomrule
    \end{tabular}
\end{table}

\paragraph{Generalization scenarios}
Beyond flat running, we evaluate transfer without changing the proposal parameters (Fig.~\ref{fig:locomotion-scenes}, right). Terrain transfer uses \emph{rocky} and \emph{mixed} tracks, with matched terrain realizations across sampling methods for a given seed. Task transfer uses box pushing and sled towing. Morphology transfer uses \emph{same-side} and \emph{diagonal} leg-length patterns with short- and long-leg scales of $0.75$ and $1.25$, respectively; the plant and planning models share the modified morphology. Planner-model mismatch is studied with a nominal plant retaining its source integrator while prediction uses either a one-step fast-RK4 planner without sub-integration or an ImplicitFast planner. These scenarios test whether proposal properties transfer across contacts, tasks, morphology, and numerical planning models. Pink is included in the transfer comparisons but is treated descriptively in the statistical analysis.

\subsubsection{Manipulation}
\begin{figure*}[!t]
    \centering
    \includegraphics[width=\textwidth]{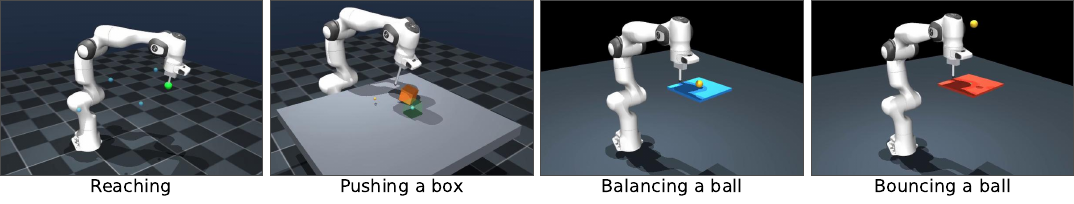}
    \caption{Representative FR3 manipulation tasks: target reaching, box pushing, ball balancing, and ball bouncing. All methods sample perturbations in three-dimensional end-effector position space.}
    \label{fig:fr3-tasks}
\end{figure*}

We evaluate the FR3 on four tasks with distinct objectives (Fig.~\ref{fig:fr3-tasks}).

\paragraph{Reaching}
The robot follows five fixed Cartesian targets in free space, traversed in the sequence $1$--$2$--$3$--$4$--$5$--$4$--$3$--$2$--$1$. With $d_t$ the fingertip-to-target distance and $s_0=\max(d_0,0.02\,\mathrm m)$, the rollout cost is
\begin{equation}
    J_{\mathrm{reach}}=\frac{1}{H}\sum_{t=1}^{H}\frac{d_t}{s_0}+2\frac{d_H}{s_0}
      +\frac{10^{-4}}{7}\sum_{t=1}^{H}\big\|(\mathbf u_t-\overline{\mathbf u}_t)\oslash\boldsymbol\sigma_u\big\|_2^2,
    \label{eq:fr3-reach-cost}
\end{equation}
where $\boldsymbol\sigma_u$ is $8\%$ of each joint range. A target is validated when the fingertip remains within $0.025\,\mathrm{m}$ of the target continuously for $1.0\,\mathrm{s}$.

\paragraph{Box pushing}
The robot pushes a box across a tabletop between predefined waypoints. The nominal task cost combines box-center distance to the active goal and fingertip distance to the closest box surface:
\begin{equation}
    J_{\mathrm{push}}=\frac{1}{H}\sum_{t=1}^{H}
      \left(10\,d_{\mathrm{box,goal}}(t)
                 +0.1\,d_{\mathrm{tip,box}}(t)\right).
    \label{eq:fr3-push-cost}
\end{equation}
The first term rewards box displacement toward the goal; the second encourages contact. Waypoints advance once the box center remains within $0.025\,\mathrm{m}$ of the active goal for $1.0\,\mathrm{s}$.

\paragraph{Ball balancing}
The robot controls a horizontal racket to balance a freely moving ball. Following ball-on-plate MPC formulations that penalize ball-position errors and control variations \cite{zarzycki2021ballplate}, as well as formulations incorporating ball velocity and plate inclination \cite{richter2011ballplate}, the objective primarily minimizes the ball's in-plane position and velocity errors. We additionally introduce penalties for racket tilt, excessive ball--racket separation, and loss of contact, together with terminal position and velocity costs. Small command-increment and control-deviation penalties are also included for stabilization. The ball-position and velocity terms are normalized using a racket radius of $0.092\,\mathrm{m}$ and a reference velocity of $0.30\,\mathrm{m/s}$, respectively.

\paragraph{Ball bouncing}
The robot repeatedly bounces a ball with the racket toward a target apex $0.20\,\mathrm{m}$ above the racket. Following model-predictive approaches to robotic ball bouncing that optimize interception position and impact velocity to achieve a desired trajectory \cite{kulchenko2011firstexit}, and hybrid MPC formulations that account for impact conditions \cite{nguyen2013juggling}, the objective primarily penalizes predicted landing-position error, deviations from the required pre-impact racket velocity, outgoing ball-velocity error, and apex-height error. The desired impact velocity is computed using a restitution coefficient of $0.56$. We introduce additional penalties for racket tilt, lateral drift, command increments and deviations, and missed contacts, while rewarding predicted successful impacts. The objective is evaluated at the $5\,\mathrm{ms}$ physics timestep, and successful task completion requires $50$ consecutive bounces without losing the ball.

The complete cost definitions, normalization factors, and numerical weights for both tasks will be made publicly available with the implementation in a GitHub repository upon publication.

\subsection{Experimental Protocol}
\label{sec:experimental-protocol}

\subsubsection{Planar ablation study}
\label{sec:ablation-protocol}
To isolate the contribution of individual motor-pool mechanisms, we use the bounded two-dimensional integrator of Fig.~\ref{fig:noise-exploration}, $\mathbf x_{t+1}=\operatorname{clip}(\mathbf x_t+\Delta t\,\mathbf u_t,-1,1)$, with $\Delta t=20\,\mathrm{ms}$, $\|\mathbf u_t\|_\infty\le1$, and the following proposal parameters $\sigma=0.3$, $L=6$, $\lambda_0=8\,\mathrm{Hz}$. In addition to the six samplers, six ablations of Spike each switch off one model component while retaining the others. Several of these interventions change more than one statistical property: \emph{w/o common drive} gives every unit an independent drive process with the statistics of Eq.~\eqref{eq:spike-common-drive}, which removes population-level synchronization but also alters the proposal spectrum; \emph{w/o size principle} gives all units the mean recruitment threshold and equal strength, a combined recruitment and strength ablation; \emph{w/o hysteresis} sets $\theta^{\mathrm{off}}_\ell=\theta^{\mathrm{on}}_\ell$; \emph{w/o rate coding} sets $f_{\max}=f_{\min}$, which also lowers the mean discharge rate; \emph{Poisson firing} replaces the renewal discharge of Eq.~\eqref{eq:spike-renewal-discharge} by memoryless Poisson discharge, removing both firing regularity and refractoriness; and \emph{uniform twitch} gives all units the base twitch kernel, which removes twitch heterogeneity and slightly shortens the mean twitch decay. Each variant is rescaled to the common pre-clipping RMS exploration magnitude.

In the open-loop experiment, the proposals directly drive the integrator from the origin for $12\,\mathrm{s}$, as in Fig.~\ref{fig:noise-exploration}. We report the proposal roughness $R_\epsilon=\frac{1}{T-1}\sum_{t=1}^{T-1}\|\boldsymbol\epsilon_t-\boldsymbol\epsilon_{t-1}\|_2$, the exploration reach $D_{\mathrm{explore}}=\big(\frac1T\sum_{t=1}^{T}\|\mathbf x_t-\mathbf x_0\|_2^2\big)^{1/2}$ over the $T$ simulated steps, and the coverage, defined as the fraction of cells of a $20\times20$ grid visited by a trajectory. In the closed-loop experiment, MPPI ($N=32$, $H=75$, LBPS with $\delta=0.95$) steers the integrator from $(-0.8,0)$ to $(0.8,0)$ around a circular obstacle of radius $0.25$ centered at the origin, with cost
\begin{equation}
    \begin{aligned}
    J={}&\sum_{t=0}^{H-1}\Big(w_g\|\mathbf x_t-\mathbf x_{\mathrm{goal}}\|_2^2
      +w_o\,\mathbb 1[\mathbf x_t\in\mathcal O]\Big)\\
      &+w_f\|\mathbf x_H-\mathbf x_{\mathrm{goal}}\|_2^2,
    \end{aligned}
    \label{eq:ablation-cost}
\end{equation}
where $w_g=1$, $w_o=10^3$, and $w_f=10$. No smoothness penalty is included. A run succeeds if the goal is reached within $0.05$ in at most $6\,\mathrm{s}$ without collision. We report the success rate, the time to goal, and the executed-control smoothness $S_u=\frac{1}{T-1}\sum_{t=1}^{T-1}\|\mathbf u_t-\mathbf u_{t-1}\|_2$. Each method is evaluated on 1000 open-loop trajectories and 1000 seeded closed-loop runs. Every method is compared with full Spike using two-sided permutation tests ($10^5$ samples): label permutation for the independent open-loop trajectories and seed-paired sign flips for the closed-loop runs. No multiple-comparison correction is applied in this ablation; its tests are reported per comparison and should be read descriptively.

Because the twitch kernels are causal and all motor units start inactive at every control update, the event-based proposals are nonstationary at the start of the horizon, whose first action is the only one executed: the first-step variance is $11\%$ (Spike) and $29\%$ (Poisson) of its mid-horizon value, versus $96$--$98\%$ for the Gaussian proposals. Three controls address this. Spike \emph{with burn-in} discards a $0.5\,\mathrm{s}$ stationary prehistory (first-step variance $99\%$); it is not used elsewhere, as it introduces a fictitious event history unrelated to the executed commands. \emph{Twitch-Gaussian} filters white Gaussian noise through the Poisson twitch from zero prehistory, sharing the spectrum and start transient of Poisson. \emph{Gaussian-Spike (full cov.)} is a Gaussian with the full $H\times H$ covariance of Spike, which reproduces all its nonstationary second-order statistics but none of its non-Gaussian structure. The last two are diagnostic constructions rather than competing designs, as they can only be derived from the proposal they imitate; the other experiments therefore retain the stationary spectrum-matched controls, which test whether frequency content alone, as exploited by colored-noise sampling, explains the observed effects.

\subsubsection{Locomotion}
The controller runs at $\Delta t=20\,\mathrm{ms}$ with horizon $H=75$, corresponding to $1.5\,\mathrm{s}$. The exploration standard deviation is 30\% of each actuator's control scale, and LBPS uses $\delta=0.95$. Poisson and Spike use $L=6$ recruitment/motor-unit levels and base rate $\lambda_0=8\,\mathrm{Hz}$. The common Poisson twitch has $\tau_r=16\,\mathrm{ms}$ and $\tau_d=64\,\mathrm{ms}$, with $200\,\mathrm{ms}$ support. Spike uses the same base twitch constants before unit-specific scaling, with $\sigma_d=0.20$ and $\tau_d^{\mathrm{drive}}=50\,\mathrm{ms}$.

\paragraph{Rollout-budget conditions}
On flat running, each sampling method is evaluated with three budgets for both robots: \emph{32} evaluates $N=32$ full-fidelity trajectories; \emph{256} evaluates $N=256$ full-fidelity trajectories; and \emph{Screen 256} draws 256 proposals, ranks them using a cheaper full-horizon ImplicitFast preview, and evaluates 32 trajectories at full fidelity. Of these, 24 are chosen by preview cost and eight are drawn uniformly from the remaining pool. Screening modifies candidate allocation, not the proposal distribution or the subsequent MPPI update. The terrain, task, morphology, and integrator-mismatch experiments use the unscreened $N=32$ condition. Pink is reported descriptively in these comparisons.

Gaussian-Poisson and Gaussian-Spike use $M_{\mathrm{cal}}=4096$ reference perturbations per proposal. Spectral factors are fitted before online control and reused during the relevant evaluations. Experiments were executed on an AMD Ryzen 7 260 processor (eight physical cores, two hardware threads per core) using Numba-accelerated numerical routines.

\paragraph{Repeated trials}
Each flat-running configuration is evaluated with ten matched seeds. For each seed, the controller executes one warm-up lap followed by ten retained laps, without resetting the controller or plant between laps. Data from the warm-up lap are discarded, and each lap has a safety limit of 2000 control steps. The seed/run, not the individual lap, is the independent statistical replicate; laps are repeated observations within a seed. Execution order is rotated deterministically across seeds to reduce timing drift. A seed is successful only if it completes the warm-up and all ten retained laps without leaving the track, falling, or exceeding the step limit. In the main flat-running runs, all seeds complete the retained evaluation. For transfer scenarios, descriptive statistics retain completed laps even when a seed later terminates early, and the completed-lap count is reported separately.

\begin{figure*}[!t]
    \centering
    \includegraphics[width=\textwidth]{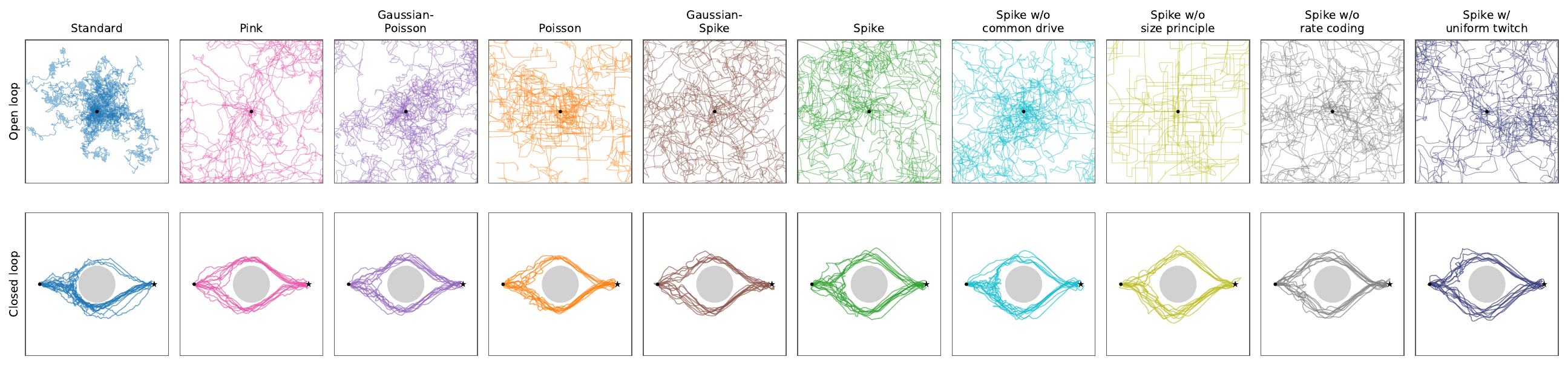}
    \caption{Planar ablation in a bounded 2D integrator. Top: 16 open-loop exploration trajectories from the origin (dot). Bottom: 16 closed-loop MPPI trajectories from the start (dot) to the goal (star) around an obstacle (grey). W/o hysteresis and Poisson firing are shown only in Table~\ref{tab:spike-ablation}.}
    \label{fig:spike-ablation}
\end{figure*}

\paragraph{Evaluation metrics}
Locomotion/task performance is summarized by the velocity of the primary body. Executed-control smoothness is the mean $\|\mathbf u_t-\mathbf u_{t-1}\|_2$ within each retained lap. Computation time per executed control step includes candidate generation, MPPI evaluation, plant integration, and progress calculations.

For Ant-2, we quantify simultaneous and differential activation. For joint $j$ at time $t$, define
\begin{equation}
    \mathrm{CI}_{t,j}=
    \begin{cases}
      \displaystyle\frac{2\min(a^+_{t,j},a^-_{t,j})}
                     {a^+_{t,j}+a^-_{t,j}}, & a^+_{t,j}+a^-_{t,j}>0,\\
      0, & a^+_{t,j}+a^-_{t,j}=0,
    \end{cases}
    \label{eq:cocontraction-index}
\end{equation}
and
\begin{equation}
    D_{t,j}=|a^+_{t,j}-a^-_{t,j}|.
    \label{eq:differential-action}
\end{equation}
Both are averaged across joints. Co-contraction measures simultaneous antagonist activation; differential action measures directional use of the antagonistic pair. Neither is optimized directly by the task objective.

Temporal coordination is quantified from executed joint commands. For Ant-2, the 16 actuator commands are reduced to eight signed joint-level signals $u_j^{\mathrm{diff}}=a_j^+-a_j^-$. For demeaned signals $x(t)$ and $y(t)$, let $\phi_x(t)$ and $\phi_y(t)$ be instantaneous phases of their analytic signals. The phase-locking value is
\begin{equation}
    C(x,y)=\left|\frac{1}{T}\sum_{t=1}^{T}
       \exp\!\left[\mathrm i(\phi_x(t)-\phi_y(t))\right]\right|,
    \label{eq:phase-locking}
\end{equation}
where $C\in[0,1]$. Within-leg coordination averages hip--ankle phase locking across four legs; between-leg coordination averages corresponding hip and ankle joint pairs over all six leg pairs.

\paragraph{Data analysis}
Distributions and descriptive statistics are reported over completed retained laps. Weighted rollout cost and computation time are averaged over controller steps within each lap; control smoothness excludes the transition between successive laps. Ant-2 co-contraction and differential action are calculated within each lap. To avoid pseudoreplication in significance testing, seed is the independent unit and signs are flipped jointly for all retained-lap differences within each seed.

We test four prespecified mechanistic contrasts: (H1) \emph{Structured sampling}, Spike versus Standard; (H2) \emph{Spectral shaping}, Gaussian-Spike versus Standard; (H3) \emph{Beyond spectral matching}, Spike versus Gaussian-Spike; and (H4) \emph{Motor-pool structure}, Spike versus Poisson. Gaussian-Poisson and Pink remain descriptive comparison methods and are excluded from these four hypothesis tests.

For each contrast, paired lap-level differences are formed only where seed and retained-lap index match. A two-sided exact seed-blocked sign-flip permutation test uses the absolute mean paired difference as its statistic. Family-wise error is controlled separately for each outcome and hypothesis using the Holm step-down procedure across robot models and rollout conditions. A metric defined on both robots has six tests per family (two models and three rollout conditions); an Ant-2-specific metric has three. Within-leg and between-leg coordination are separate outcomes. Raw and Holm-adjusted $p$-values are denoted $p$ and $p_{\mathrm H}$; significance is assessed at $\alpha=0.05$. Descriptive effect magnitudes accompany inferential results.

\subsubsection{Manipulation}
All FR3 manipulation tasks use a control period of $\Delta t=10\,\mathrm{ms}$, with a MuJoCo physics timestep of $5\,\mathrm{ms}$ and the implicit integrator configured in the supplied simulation scenes. Exploration is generated in three Cartesian proposal coordinates. Table~\ref{tab:fr3-settings} lists the task-specific settings. The nominal plan is included as one of the $N$ rollouts. Each manipulation task is evaluated over 100 independently seeded runs.

\begin{table}[!htbp]
\centering
\caption{FR3 controller settings. All tasks use $N=32$, $\Delta t=10\,\mathrm{ms}$, LBPS with $\delta=0.95$, and warm starting. Commands are clipped to the joint range and rate-limited to $|\Delta q_k|\le\dot q_k^{\max}\Delta t$, with $\dot q^{\max}=2.175\,\mathrm{rad/s}$ for joints 1--4 and $2.61\,\mathrm{rad/s}$ for joints 5--7, in both rollouts and plant.}
\label{tab:fr3-settings}
\scriptsize
\setlength{\tabcolsep}{3pt}
\begin{tabular}{lcccl}
\toprule
Task & $H$ & $\sigma$ (m) & Limit (s) & Success \\
\midrule
Reaching & 20 & $0.01\,s(d)^{\dagger}$ & 40 & 9 targets, $2.5\,\mathrm{cm}$ for $1\,\mathrm s$ \\
Pushing & 100 & 0.005 & 20 & 3 waypoints, $2.5\,\mathrm{cm}$ for $1\,\mathrm s$ \\
Balancing & 20 & 0.01 & 20 & ball kept on the plate \\
Bouncing & 100 & 0.01 & 25 & 50 bounces \\
\bottomrule
\multicolumn{5}{l}{$^{\dagger}s(d)=0.1+0.9\min(1,d/0.5\,\mathrm m)$, with $d$ the tip--target distance.}
\end{tabular}
\end{table}

\begin{table}[t]
\centering
\caption{Spike ablation in a bounded 2D integrator. Open loop: proposal roughness $R_\epsilon$, exploration reach $D_{\mathrm{explore}}$ and grid coverage over 1000 trajectories. Closed loop: success rate, time to goal (successful runs) and executed-control smoothness $S_u$ over 1000 MPPI runs. Mean $\pm$ SD. \colorbox{best}{Green}: three best; \colorbox{worst}{red}: three worst per column.}
\label{tab:spike-ablation}
\scriptsize
\setlength{\tabcolsep}{2.5pt}
\renewcommand{\arraystretch}{1.05}

\resizebox{\columnwidth}{!}{%
\begin{tabular}{lccc}
\toprule
\multicolumn{4}{c}{\textit{Open loop}} \\
\midrule
Sampler & $R_\epsilon$ ($\times10^{-3}$) $\downarrow$ & $D_{\mathrm{explore}}$ $\uparrow$ & Coverage (\%) $\uparrow$ \\
\midrule
Standard & \cellcolor{worst}460.1 $\pm$ 11.1 & \cellcolor{worst}0.433 $\pm$ 0.160 & \cellcolor{worst}8.4 $\pm$ 1.5 \\
Pink & \cellcolor{worst}262.0 $\pm$ 5.8 & \cellcolor{best}0.978 $\pm$ 0.124 & 15.8 $\pm$ 3.6 \\
Gaussian-Poisson & \cellcolor{worst}213.3 $\pm$ 4.9 & \cellcolor{worst}0.712 $\pm$ 0.168 & 16.0 $\pm$ 2.6 \\
Twitch-Gaussian & 212.9 $\pm$ 4.9 & 0.714 $\pm$ 0.170 & 15.9 $\pm$ 2.5 \\
Poisson & 145.4 $\pm$ 11.0 & \cellcolor{worst}0.712 $\pm$ 0.175 & \cellcolor{worst}14.5 $\pm$ 2.4 \\
Gaussian-Spike & 134.3 $\pm$ 3.5 & \cellcolor{best}0.785 $\pm$ 0.148 & \cellcolor{best}18.1 $\pm$ 3.0 \\
Gaussian-Spike (full cov.) & 133.8 $\pm$ 3.6 & 0.779 $\pm$ 0.153 & \cellcolor{best}18.2 $\pm$ 3.1 \\
Spike (full) & 114.6 $\pm$ 6.5 & 0.779 $\pm$ 0.149 & 17.4 $\pm$ 3.0 \\
\midrule
\quad with burn-in & 114.1 $\pm$ 6.3 & 0.778 $\pm$ 0.154 & 17.5 $\pm$ 2.9 \\
\quad w/o common drive & 157.0 $\pm$ 7.0 & 0.754 $\pm$ 0.160 & 16.5 $\pm$ 2.7 \\
\quad w/o size principle & \cellcolor{best}87.9 $\pm$ 12.7 & 0.783 $\pm$ 0.167 & \cellcolor{worst}12.3 $\pm$ 2.5 \\
\quad w/o hysteresis & \cellcolor{best}113.4 $\pm$ 6.5 & 0.779 $\pm$ 0.153 & 16.9 $\pm$ 2.9 \\
\quad w/o rate coding & 119.6 $\pm$ 6.6 & 0.777 $\pm$ 0.152 & \cellcolor{best}17.5 $\pm$ 3.0 \\
\quad Poisson firing & \cellcolor{best}113.6 $\pm$ 5.9 & \cellcolor{best}0.783 $\pm$ 0.152 & 17.2 $\pm$ 3.0 \\
\quad uniform twitch & 127.9 $\pm$ 7.0 & 0.763 $\pm$ 0.156 & 16.8 $\pm$ 2.9 \\
\midrule
\multicolumn{4}{c}{\textit{Closed loop}} \\
\midrule
Sampler & Success (\%) $\uparrow$ & Time (s) $\downarrow$ & $S_u$ ($\times10^{-3}$) $\downarrow$ \\
\midrule
Standard & 94.7 & \cellcolor{worst}3.77 $\pm$ 0.72 & \cellcolor{worst}572.2 $\pm$ 32.5 \\
Pink & 100.0 & 2.77 $\pm$ 0.43 & \cellcolor{worst}345.6 $\pm$ 27.7 \\
Gaussian-Poisson & 100.0 & 2.79 $\pm$ 0.42 & \cellcolor{worst}260.5 $\pm$ 18.9 \\
Twitch-Gaussian & 99.9 & 2.84 $\pm$ 0.56 & 246.3 $\pm$ 20.0 \\
Poisson & 99.8 & \cellcolor{worst}2.93 $\pm$ 0.54 & 214.6 $\pm$ 17.3 \\
Gaussian-Spike & 100.0 & \cellcolor{best}2.66 $\pm$ 0.37 & 185.7 $\pm$ 13.9 \\
Gaussian-Spike (full cov.) & 99.9 & \cellcolor{best}2.63 $\pm$ 0.45 & 153.3 $\pm$ 14.0 \\
Spike (full) & 100.0 & 2.68 $\pm$ 0.45 & \cellcolor{best}142.8 $\pm$ 13.1 \\
\midrule
\quad with burn-in & 100.0 & 2.72 $\pm$ 0.47 & 166.8 $\pm$ 13.1 \\
\quad w/o common drive & 99.9 & 2.79 $\pm$ 0.52 & 200.4 $\pm$ 17.8 \\
\quad w/o size principle & 99.7 & \cellcolor{worst}2.98 $\pm$ 0.61 & \cellcolor{best}141.0 $\pm$ 12.7 \\
\quad w/o hysteresis & 100.0 & \cellcolor{best}2.67 $\pm$ 0.48 & 145.7 $\pm$ 13.4 \\
\quad w/o rate coding & 100.0 & 2.68 $\pm$ 0.46 & 151.4 $\pm$ 14.4 \\
\quad Poisson firing & 100.0 & 2.68 $\pm$ 0.46 & \cellcolor{best}139.7 $\pm$ 12.9 \\
\quad uniform twitch & 100.0 & 2.74 $\pm$ 0.51 & 159.1 $\pm$ 14.9 \\
\bottomrule
\end{tabular}%
}
\end{table}

\subsubsection{Real-robot experiments}
We additionally evaluate reaching on a physical FR3. The high-level MPPI controller predicts joint-position-controlled motion in MuJoCo at $100\,\mathrm{Hz}$, using the configured implicit integrator. The desired joint positions are transmitted over Ethernet UDP at $100\,\mathrm{Hz}$. On the robot-side workstation, a ROS~2 interface runs at $1\,\mathrm{kHz}$ and implements proportional position tracking,
\begin{equation}
    \dot{\mathbf q}_{\mathrm{cmd}}=
        K_p(\mathbf q_{\mathrm{des}}-\mathbf q),
    \qquad K_p=4.0\,\mathrm{s}^{-1},
    \label{eq:fr3-hardware-proportional}
\end{equation}
where $\mathbf q_{\mathrm{des}}$ and $\mathbf q$ are desired and measured joint positions. The resulting desired joint-velocity commands are sent through the Franka Control Interface, which supports $1\,\mathrm{kHz}$ command and state exchange \cite{franka2026fci}. Measured joint positions and velocities are returned over UDP and used to initialize subsequent simulated MPPI rollouts.

To reduce the mismatch between the predicted and the executed motion, the simulation model includes two approximations of the hardware interface. First, the tracking delay introduced by Eq.~\eqref{eq:fr3-hardware-proportional} is reproduced in simulation. The MuJoCo position actuators are reparameterized so that, with the arm's gravity compensated, each joint follows the same first-order response. Without this change, the stiffer default servos track targets several times faster than the robot. Second, the simulator state is not only initialized from the measured positions and velocities. The actuator command is also set to the last desired joint position sent to the robot. Because the actuator force, and therefore the joint acceleration, depends on this command, this gives an approximation of the robot's current acceleration, which is not measured by the FR3 sensors. Each rollout thus starts from the motion the robot is actually executing rather than from rest at the measured configuration.

Reaching error on hardware is estimated by forward kinematics of the feedback joint configuration, without an independent Cartesian motion-tracking system. Beyond these two approximations, the interface closes the loop on measured state but does not explicitly compensate for other model--plant discrepancies or unmodeled contacts.

\paragraph{Baseline selection and objective design}
The baselines isolate the influence of the proposal distribution rather than compare MPPI formulations: Standard provides a temporally correlated Gaussian reference, Pink long-range temporal correlations, Poisson sparse event-based sampling with twitch filtering, and Gaussian-Poisson and Gaussian-Spike the second-order spectral content of the event-driven proposals. The objectives are deliberately task-driven, and no smoothness regularization is introduced to favor the proposed samplers; the small stabilization and control-deviation terms, as well as the command-variation penalty used in ball balancing, are identical across methods. Differences in smoothness therefore emerge from the proposal distributions under objectives shared by all methods.

\section{Results}

\begin{figure*}[!t]
    \centering
    \includegraphics[width=\textwidth]{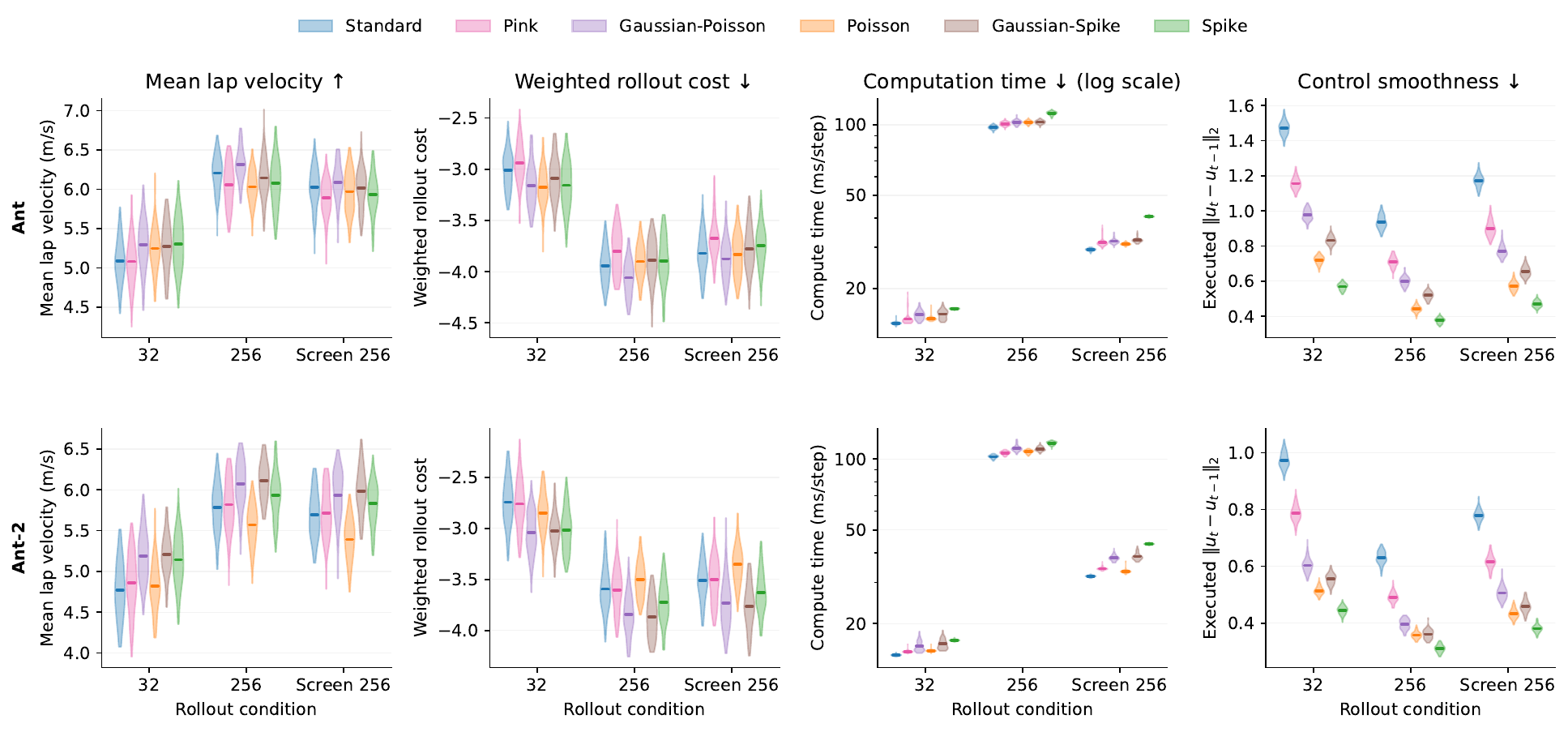}
    \caption{Flat-running performance for Ant and Ant-2 under the three rollout conditions. Panels show mean lap velocity, weighted rollout cost, computation time per control step, and executed-control smoothness $\|\mathbf u_t-\mathbf u_{t-1}\|_2$. Distributions use completed retained laps, and horizontal markers indicate means. Arrows show the preferred direction of each metric. Pink is a descriptive comparison on the flat-running task.}
    \label{fig:flat-main}
\end{figure*}

\begin{figure}[!htbp]
    \centering
    \includegraphics[width=\linewidth]{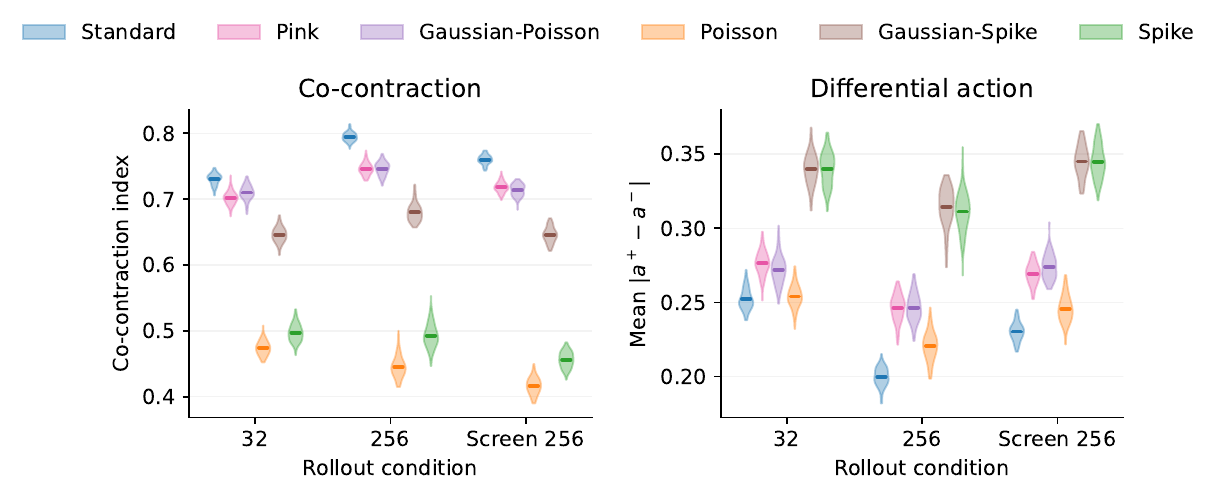}
    \caption{Ant-2 agonist--antagonist actuation on flat running across rollout conditions. The co-contraction index measures simultaneous activation; differential action quantifies net directional activation. Pink is included as a descriptive baseline.}
    \label{fig:flat-ant2-actuation}
\end{figure}

\begin{figure}[!htbp]
    \centering
    \includegraphics[width=\columnwidth]{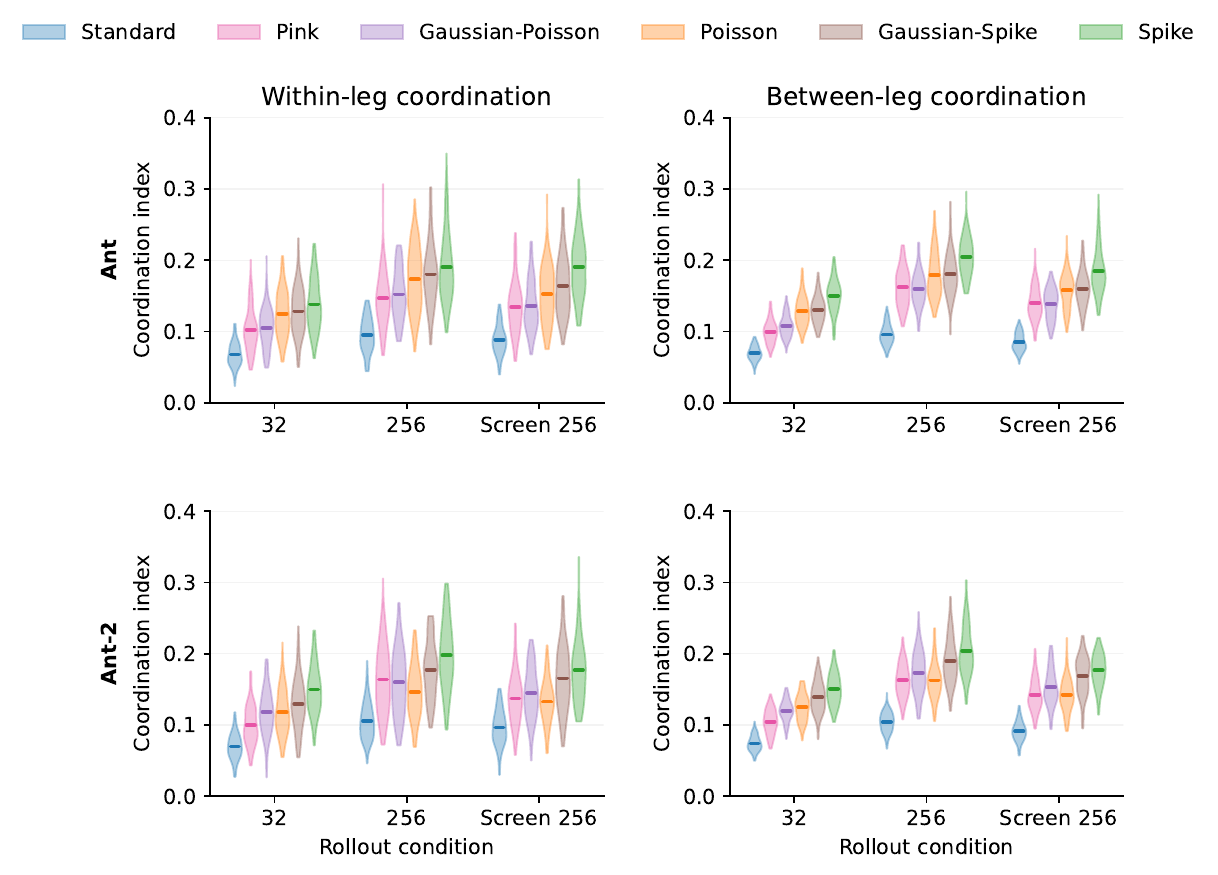}
    \caption{Executed-command coordination for Ant and Ant-2 on flat running. Within-leg coordination describes hip--ankle phase consistency; between-leg coordination describes the phase consistency of homologous joints over leg pairs. Higher values indicate more consistent relative phase.}
    \label{fig:coordination}
\end{figure}

\subsection{Ablation of Motor-Pool Components}
\label{sec:ablation-results}

Figure~\ref{fig:spike-ablation} and Table~\ref{tab:spike-ablation} report the planar ablation. The closed-loop task is solved by nearly all samplers: every method succeeds in at least $99.7\%$ of the runs except Standard ($94.7\%$, $p<10^{-4}$ against Spike). Full Spike yields smoother commands than Standard ($S_u$ reduced by $75.0\%$) and Gaussian-Spike ($23.1\%$; $p<10^{-4}$ for both), with slightly lower coverage than Gaussian-Spike ($17.4\%$ versus $18.1\%$, $p<10^{-4}$). With burn-in, $S_u$ increases by $16.8\%$ but remains $10.2$--$70.9\%$ below that of every baseline ($p<10^{-4}$), with a similar time to goal ($2.72$ versus $2.68\,\mathrm{s}$, $p=0.049$). Twitch-Gaussian is only $5.4\%$ smoother than Gaussian-Poisson and remains $14.8\%$ rougher than Poisson, whereas Gaussian-Spike (full cov.) closes about three quarters of the gap between Gaussian-Spike and Spike, which remains $6.9\%$ smoother ($p<10^{-4}$ for all); the full-covariance control nevertheless reaches the goal slightly faster ($2.63\,\mathrm{s}$, $p=0.008$). After clipping, all samplers explore at a similar level, with a horizon-averaged perturbation variance of $0.80$--$0.94\,\sigma^2$, $4$--$6\%$ of clipped commands, and an effective sample size of $18.6$--$20.6$ out of $N=32$.

The ablations show that the mechanisms contribute unequally. Removing the common drive has the largest effect: it increases proposal roughness by $37.0\%$ and executed-control roughness by $40.3\%$, reduces coverage to $16.5\%$, and lengthens the time to goal ($2.79$ versus $2.68\,\mathrm{s}$; $p<10^{-4}$ for all). It also roughly doubles the share of proposal power above $5\,\mathrm{Hz}$ ($13.4\%$ versus $6.5\%$). Replacing heterogeneous twitches with a uniform kernel increases $R_\epsilon$ by $11.6\%$ and $S_u$ by $11.4\%$ ($p<10^{-4}$) and reduces coverage to $16.8\%$ ($p<10^{-4}$). Removing rate coding has a smaller effect on smoothness ($+4.4\%$ and $+6.0\%$, $p<10^{-4}$) without affecting coverage or time to goal ($p\ge0.46$). Removing hysteresis and replacing renewal discharge by Poisson firing change $S_u$ by only $+2.0\%$ and $-2.2\%$ ($p<10^{-4}$), without changing the time to goal ($p\ge0.69$). Removing the size principle lowers proposal roughness by $23.3\%$ but $S_u$ by only $1.3\%$ ($p=0.002$). This results from synchronous recruitment around a single threshold: the pools remain silent for long periods, producing axis-aligned exploration, $29.4\%$ lower coverage, and a longer time to goal ($2.98$ versus $2.68\,\mathrm{s}$; $p<10^{-4}$ for both).

\subsection{Flat Racing Comparison}
\label{sec:results-flat}

\begin{table}[!htbp]
\centering
\caption{Significance summary for all hypotheses at the $0.05$ threshold
after Holm correction. S: Spike; GS: Gaussian-Spike; P: Poisson;
Std: Standard Gaussian. Entries correspond to $32/256/\mathrm{Scr.}$.}
\label{tab:hypothesis-summary}
\scriptsize
\setlength{\tabcolsep}{2.8pt}
\renewcommand{\arraystretch}{1.12}

\resizebox{\columnwidth}{!}{%
\begin{tabular}{llcccc}
\hline
\textbf{Outcome} &
\textbf{Robot} &
\textbf{H1: S--Std} &
\textbf{H2: GS--Std} &
\textbf{H3: S--GS} &
\textbf{H4: S--P} \\
\hline

\multicolumn{6}{l}{\textit{Task performance}} \\

Lap velocity
& Ant
& $\checkmark/\checkmark/\checkmark$
& $\checkmark/-/-$
& $-/-/-$
& $-/-/-$ \\

& Ant-2
& $\checkmark/\checkmark/\checkmark$
& $\checkmark/-/-$
& $-/-/-$
& $\checkmark/\checkmark/\checkmark$ \\[0.5mm]

Weighted rollout cost
& Ant
& $\checkmark/-/-$
& $-/-/-$
& $-/-/-$
& $-/-/-$ \\

& Ant-2
& $\checkmark/\checkmark/\checkmark$
& $\checkmark/-/-$
& $-/-/-$
& $\checkmark/\checkmark/\checkmark$ \\

\hline
\multicolumn{6}{l}{\textit{Temporal structure and coordination}} \\

Control smoothness
& Ant
& $\checkmark/\checkmark/\checkmark$
& $\checkmark/\checkmark/\checkmark$
& $\checkmark/\checkmark/\checkmark$
& $\checkmark/\checkmark/\checkmark$ \\

& Ant-2
& $\checkmark/\checkmark/\checkmark$
& $\checkmark/\checkmark/\checkmark$
& $\checkmark/\checkmark/\checkmark$
& $\checkmark/\checkmark/\checkmark$ \\[0.5mm]

Within-leg coordination
& Ant
& $\checkmark/\checkmark/\checkmark$
& $\checkmark/\checkmark/\checkmark$
& $-/-/\checkmark$
& $\checkmark/\checkmark/\checkmark$ \\

& Ant-2
& $\checkmark/\checkmark/\checkmark$
& $\checkmark/\checkmark/\checkmark$
& $\checkmark/-/-$
& $\checkmark/\checkmark/\checkmark$ \\[0.5mm]

Between-leg coordination
& Ant
& $\checkmark/\checkmark/\checkmark$
& $\checkmark/\checkmark/\checkmark$
& $\checkmark/-/\checkmark$
& $\checkmark/\checkmark/\checkmark$ \\

& Ant-2
& $\checkmark/\checkmark/\checkmark$
& $\checkmark/\checkmark/\checkmark$
& $-/-/-$
& $\checkmark/\checkmark/\checkmark$ \\

\hline
\multicolumn{6}{l}{\textit{Antagonistic actuation}} \\

Co-contraction
& Ant-2
& $\checkmark/\checkmark/\checkmark$
& $\checkmark/\checkmark/\checkmark$
& $\checkmark/\checkmark/\checkmark$
& $\checkmark/\checkmark/\checkmark$ \\[0.5mm]

Differential action
& Ant-2
& $\checkmark/\checkmark/\checkmark$
& $\checkmark/\checkmark/\checkmark$
& $-/-/-$
& $\checkmark/\checkmark/\checkmark$ \\

\hline
\end{tabular}%
}
\end{table}

\begin{table}[!htbp]
\centering
\caption{Generalization to mixed and rocky terrain. Velocity and
smoothness are reported as mean $\pm$ standard deviation over completed
laps.}
\label{tab:terrain}
\scriptsize
\setlength{\tabcolsep}{2.5pt}
\renewcommand{\arraystretch}{1.05}

\resizebox{\columnwidth}{!}{%
\begin{tabular}{lllccc}
\toprule
Terrain & Robot & Sampler
& Vel. (m/s) $\uparrow$
& Laps/seed $\uparrow$
& Smooth. $\downarrow$ \\
\midrule

\multirow{12}{*}{Mixed}
& \multirow{6}{*}{Ant}
& Standard         & $3.270 \pm 1.102$ & 5.7/10 & $1.451 \pm 0.128$ \\
& & Pink             & $3.677 \pm 1.263$ & 4.8/10 & $1.142 \pm 0.105$ \\
& & Gaussian-Poisson & $3.632 \pm 1.231$ & \textbf{10.0/10} & $0.962 \pm 0.091$ \\
& & Poisson          & $3.992 \pm 1.199$ & 9.5/10 & $0.734 \pm 0.068$ \\
& & Gaussian-Spike   & $3.903 \pm 1.160$ & 8.0/10 & $0.815 \pm 0.072$ \\
& & Spike            & $\mathbf{4.186 \pm 1.121}$ & 8.8/10 & $\mathbf{0.574 \pm 0.048}$ \\
\cmidrule(lr){2-6}

& \multirow{6}{*}{Ant-2}
& Standard         & $3.831 \pm 0.397$ & 1.6/10 & $1.098 \pm 0.038$ \\
& & Pink             & $\mathbf{4.195 \pm 0.481}$ & 0.8/10 & $0.839 \pm 0.022$ \\
& & Gaussian-Poisson & $4.037 \pm 1.070$ & 5.8/10 & $0.645 \pm 0.052$ \\
& & Poisson          & $3.264 \pm 1.105$ & 3.1/10 & $0.533 \pm 0.059$ \\
& & Gaussian-Spike   & $4.085 \pm 1.032$ & \textbf{8.5/10} & $0.568 \pm 0.048$ \\
& & Spike            & $4.110 \pm 0.938$ & 8.0/10 & $\mathbf{0.459 \pm 0.038}$ \\
\midrule

\multirow{12}{*}{Rocky}
& \multirow{6}{*}{Ant}
& Standard         & $4.154 \pm 0.698$ & 2.2/10 & $1.637 \pm 0.051$ \\
& & Pink             & $3.923 \pm 1.049$ & \textbf{4.1/10} & $1.247 \pm 0.075$ \\
& & Gaussian-Poisson & $4.333 \pm 0.786$ & 3.5/10 & $1.048 \pm 0.052$ \\
& & Poisson          & $4.402 \pm 0.887$ & 3.8/10 & $0.784 \pm 0.041$ \\
& & Gaussian-Spike   & $4.480 \pm 0.745$ & 4.0/10 & $0.891 \pm 0.049$ \\
& & Spike            & $\mathbf{4.656 \pm 0.768}$ & 4.0/10 & $\mathbf{0.605 \pm 0.029}$ \\
\cmidrule(lr){2-6}

& \multirow{6}{*}{Ant-2}
& Standard         & $3.861 \pm 0.977$ & 2.4/10 & $1.152 \pm 0.086$ \\
& & Pink             & $4.173 \pm 0.786$ & \textbf{4.1/10} & $0.887 \pm 0.049$ \\
& & Gaussian-Poisson & $4.161 \pm 0.976$ & 4.0/10 & $0.689 \pm 0.042$ \\
& & Poisson          & $3.738 \pm 0.988$ & 3.5/10 & $0.568 \pm 0.038$ \\
& & Gaussian-Spike   & $4.219 \pm 0.996$ & 2.5/10 & $0.607 \pm 0.048$ \\
& & Spike            & $\mathbf{4.477 \pm 0.861}$ & \textbf{4.1/10} & $\mathbf{0.491 \pm 0.031}$ \\
\bottomrule
\end{tabular}%
}
\end{table}

\begin{table}[!htbp]
\centering
\caption{Generalization to object-interaction tasks.}
\label{tab:tasks}
\scriptsize
\setlength{\tabcolsep}{2.5pt}
\renewcommand{\arraystretch}{1.05}

\resizebox{\columnwidth}{!}{%
\begin{tabular}{lllccc}
\toprule
Task & Robot & Sampler
& Vel. (m/s) $\uparrow$
& Laps/seed $\uparrow$
& Smooth. $\downarrow$ \\
\midrule

\multirow{12}{*}{Push box}
& \multirow{6}{*}{Ant}
& Standard           & $\mathbf{2.324 \pm 0.218}$ & \textbf{10.0/10} & $1.923 \pm 0.033$ \\
& & Pink             & $2.231 \pm 0.215$ & 9.0/10 & $1.502 \pm 0.026$ \\
& & Gaussian-Poisson & $2.008 \pm 0.194$ & 9.0/10 & $1.278 \pm 0.022$ \\
& & Poisson          & $2.176 \pm 0.183$ & 9.1/10 & $1.026 \pm 0.021$ \\
& & Gaussian-Spike   & $2.082 \pm 0.218$ & \textbf{10.0/10} & $1.075 \pm 0.020$ \\
& & Spike            & $2.219 \pm 0.240$ & \textbf{10.0/10} & $\mathbf{0.748 \pm 0.015}$ \\
\cmidrule(lr){2-6}

& \multirow{6}{*}{Ant-2}
& Standard           & $2.372 \pm 0.281$ & \textbf{10.0/10} & $1.287 \pm 0.020$ \\
& & Pink             & $2.413 \pm 0.259$ & \textbf{10.0/10} & $1.029 \pm 0.016$ \\
& & Gaussian-Poisson & $\mathbf{2.449 \pm 0.244}$ & \textbf{10.0/10} & $0.854 \pm 0.017$ \\
& & Poisson          & $2.337 \pm 0.224$ & 9.7/10 & $0.701 \pm 0.014$ \\
& & Gaussian-Spike   & $2.344 \pm 0.229$ & 9.4/10 & $0.756 \pm 0.016$ \\
& & Spike            & $2.429 \pm 0.228$ & \textbf{10.0/10} & $\mathbf{0.594 \pm 0.010}$ \\
\midrule

\multirow{12}{*}{Tow sled}
& \multirow{6}{*}{Ant}
& Standard           & $2.650 \pm 0.473$ & 8.5/10 & $1.605 \pm 0.098$ \\
& & Pink             & $\mathbf{2.652 \pm 0.424}$ & \textbf{10.0/10} & $1.218 \pm 0.084$ \\
& & Gaussian-Poisson & $2.580 \pm 0.543$ & 9.0/10 & $0.986 \pm 0.088$ \\
& & Poisson          & $2.399 \pm 0.608$ & 9.2/10 & $0.721 \pm 0.081$ \\
& & Gaussian-Spike   & $2.491 \pm 0.583$ & 7.6/10 & $0.820 \pm 0.087$ \\
& & Spike            & $2.647 \pm 0.500$ & 8.4/10 & $\mathbf{0.577 \pm 0.044}$ \\
\cmidrule(lr){2-6}

& \multirow{6}{*}{Ant-2}
& Standard           & $2.540 \pm 0.402$ & \textbf{10.0/10} & $1.152 \pm 0.045$ \\
& & Pink             & $2.609 \pm 0.385$ & 7.7/10 & $0.900 \pm 0.043$ \\
& & Gaussian-Poisson & $2.708 \pm 0.467$ & 9.3/10 & $0.693 \pm 0.051$ \\
& & Poisson          & $2.539 \pm 0.415$ & \textbf{10.0/10} & $0.585 \pm 0.042$ \\
& & Gaussian-Spike   & $\mathbf{2.885 \pm 0.396}$ & 9.6/10 & $0.610 \pm 0.041$ \\
& & Spike            & $2.680 \pm 0.536$ & 9.8/10 & $\mathbf{0.470 \pm 0.044}$ \\
\bottomrule
\end{tabular}%
}
\end{table}

Figure~\ref{fig:flat-main} compares the sampling methods on the flat racing task for both robot morphologies. Table~\ref{tab:hypothesis-summary} summarizes the significance of the four prespecified contrasts after Holm correction. Unless otherwise stated, reported $p_{\mathrm H}$ values are Holm-adjusted values from the seed-blocked permutation tests described in Sec.~\ref{sec:experimental-protocol}.

At the low rollout budget ($N=32$), both structured proposals numerically improve velocity on the torque-driven Ant compared with Standard Gaussian sampling. Mean lap velocity increases from $5.09~\mathrm{m/s}$ with Standard to $5.25~\mathrm{m/s}$ with Poisson and $5.30~\mathrm{m/s}$ with Spike, corresponding to increases of $3.1\%$ and $4.2\%$, respectively. The Spike--Standard difference is significant ($p_{\mathrm H}=0.016$).

The effect is more pronounced on the antagonistic Ant-2. At $N=32$, Poisson produces only a small velocity increase from $4.77$ to $4.82~\mathrm{m/s}$, whereas Spike reaches $5.14~\mathrm{m/s}$, a $7.8\%$ increase over Standard ($p_{\mathrm H}=0.012$). Spike also significantly outperforms Poisson in velocity ($p_{\mathrm H}=0.020$) and yields the lowest weighted rollout cost ($-3.01$), compared with $-2.85$ for Poisson and $-2.74$ for Standard (both contrasts $p_{\mathrm H}\leq0.020$).

Increasing the full rollout budget to $N=256$ changes the relative performance on the two morphologies. On Ant, Standard reaches $6.21~\mathrm{m/s}$, compared with $6.03~\mathrm{m/s}$ for Poisson and $6.08~\mathrm{m/s}$ for Spike, with Spike significantly slower than Standard ($p_{\mathrm H}=0.029$). On Ant-2, Spike instead reaches $5.93~\mathrm{m/s}$ compared with $5.78~\mathrm{m/s}$ for Standard ($p_{\mathrm H}=0.012$), whereas Poisson decreases the mean velocity to $5.57~\mathrm{m/s}$. Spike also produces the lowest weighted rollout cost among these three methods.

With rollout screening, Standard, Poisson, and Spike reach $6.03$, $5.97$, and $5.93~\mathrm{m/s}$ on Ant, respectively. On Ant-2, Poisson decreases velocity from $5.69$ to $5.39~\mathrm{m/s}$, whereas Spike reaches $5.83~\mathrm{m/s}$, significantly above Standard ($p_{\mathrm H}=0.031$) and Poisson ($p_{\mathrm H}=0.020$). Thus, Spike significantly improves velocity over Standard on Ant-2 across all three rollout conditions, whereas the direction of the difference depends on the rollout budget for Ant. In contrast, no significant velocity difference between Spike and Gaussian-Spike is detected in any flat-running condition ($p_{\mathrm H}\geq0.094$).

The strongest and most consistent effect of the proposal structure is observed in executed-control smoothness. Relative to Standard, Poisson reduces the mean control-increment norm by approximately $43$--$53\%$, while Spike reduces it by approximately $51$--$61\%$ across the two robots and three rollout conditions. Spike also produces smaller control increments than Poisson and Gaussian-Spike in every condition. All four prespecified contrasts are significant for smoothness across both robots and all rollout conditions ($p_{\mathrm H}\leq0.047$).

The smoother structured proposals incur additional computation: at $N=32$, Spike requires $16.4$ and $17.0~\mathrm{ms/step}$ on Ant and Ant-2, versus $14.2$--$15.3~\mathrm{ms/step}$ for Standard and Poisson ($p_{\mathrm H}=0.012$). These differences remain small compared with the $97$--$117~\mathrm{ms/step}$ required by 256 full rollouts, or $29$--$44~\mathrm{ms/step}$ with screening.

Figure~\ref{fig:flat-ant2-actuation} further shows that the proposal structure systematically changes how the antagonistic actuation space is used. Standard produces substantially greater simultaneous agonist--antagonist activation than either structured proposal. At $N=32$, the mean co-contraction index decreases from $0.73$ with Standard to $0.47$ with Poisson and $0.50$ with Spike, with similar reductions at $N=256$ and with screening. Spike differs significantly from both Standard and Poisson across all three conditions ($p_{\mathrm H}=0.006$ for both contrasts).

In contrast, Spike produces the largest differential antagonist action. At $N=32$, the mean $|a^+-a^-|$ is $0.25$ for Standard, $0.25$ for Poisson, and $0.34$ for Spike. These differences are significant for Spike relative to Standard and Poisson across all rollout conditions ($p_{\mathrm H}=0.006$), but not relative to Gaussian-Spike ($p_{\mathrm H}\geq0.656$). The proposal structure thus changes how the same antagonistic actuator model is used.

Figure~\ref{fig:coordination} finally shows that structured sampling also changes the coordination of the executed commands. For both Ant and Ant-2, Poisson numerically increases within-leg and between-leg coordination relative to Standard under all three rollout conditions, while Spike produces a further increase. The Spike--Standard and Spike--Poisson differences are significant for both coordination metrics across all conditions ($p_{\mathrm H}\leq0.039$).

The spectrum-matched Gaussian control reproduces a substantial part of this increase: only four of the twelve Spike--Gaussian-Spike coordination contrasts are significant after correction, so the additional contribution of the full Spike proposal to coordination is less consistent than its effect on control smoothness.

\subsection{Generalization and Robustness to Novel Locomotion Conditions}
\label{sec:results-generalization}

Across Tables~\ref{tab:terrain}--\ref{tab:integrator-mismatch}, velocity and control smoothness are reported as mean $\pm$ standard deviation over completed retained laps, including laps completed by seeds that subsequently terminate early. Bold values denote the best descriptive value within each scenario--robot block according to the indicated metric direction. Because velocity and smoothness are conditioned on completed laps, these metrics should be interpreted jointly with the reported completion rate, particularly in scenarios with frequent early termination.

\paragraph{Terrain generalization}
Table~\ref{tab:terrain} shows that the structured proposals retain their smooth-control behavior under substantially more difficult terrain. Spike achieves the highest mean velocity in three out of four of the terrain--morphology combinations and the lowest control increments throughout. These smoothness differences are significant ($p_\mathrm{H}\leq 0.047$) but the velocity differences are not significant for any contrast on either terrain ($p_\mathrm{H}\geq 0.063$), as few seeds complete laps under both samplers in each pair.

\paragraph{Task generalization}
The object-interaction results in Table~\ref{tab:tasks} show smaller differences in task velocity, with all methods generally retaining high completion rates. Spike nevertheless produces the smoothest control in every robot--task combination, and all four contrasts are significant for smoothness in every combination ($p_\mathrm{H}=0.008$). On Ant, Standard remains fastest for pushing ($+0.11~\mathrm{m/s}$ over Spike; $p_\mathrm{H}=0.051$), while Pink, Standard, and Spike are essentially tied for towing ($2.652$, $2.650$, and $2.647~\mathrm{m/s}$; H1, $p_\mathrm{H}=0.81$), indicating that smoother control does not consistently impose a substantial task-velocity penalty in these scenarios.

\begin{table}[!htbp]
\centering
\caption{Robustness to altered morphology.}
\label{tab:leg-mismatch}
\scriptsize
\setlength{\tabcolsep}{2.5pt}
\renewcommand{\arraystretch}{1.05}

\resizebox{\columnwidth}{!}{%
\begin{tabular}{lllccc}
\toprule
Asymmetry & Robot & Sampler
& Vel. (m/s) $\uparrow$
& Laps/seed $\uparrow$
& Smooth. $\downarrow$ \\
\midrule

\multirow{12}{*}{Diagonal}
& \multirow{6}{*}{Ant}
& Standard           & $4.764 \pm 0.351$ & \textbf{10.0/10} & $1.497 \pm 0.046$ \\
& & Pink             & $4.763 \pm 0.353$ & \textbf{10.0/10} & $1.174 \pm 0.035$ \\
& & Gaussian-Poisson & $4.956 \pm 0.359$ & \textbf{10.0/10} & $1.001 \pm 0.030$ \\
& & Poisson          & $\mathbf{4.972 \pm 0.356}$ & \textbf{10.0/10} & $0.739 \pm 0.027$ \\
& & Gaussian-Spike   & $4.886 \pm 0.349$ & \textbf{10.0/10} & $0.844 \pm 0.030$ \\
& & Spike            & $4.930 \pm 0.333$ & \textbf{10.0/10} & $\mathbf{0.584 \pm 0.020}$ \\
\cmidrule(lr){2-6}

& \multirow{6}{*}{Ant-2}
& Standard           & $4.438 \pm 0.395$ & \textbf{10.0/10} & $1.007 \pm 0.033$ \\
& & Pink             & $4.568 \pm 0.400$ & \textbf{10.0/10} & $0.806 \pm 0.026$ \\
& & Gaussian-Poisson & $4.835 \pm 0.307$ & \textbf{10.0/10} & $0.625 \pm 0.023$ \\
& & Poisson          & $4.468 \pm 0.374$ & \textbf{10.0/10} & $0.536 \pm 0.020$ \\
& & Gaussian-Spike   & $\mathbf{4.946 \pm 0.350}$ & \textbf{10.0/10} & $0.565 \pm 0.021$ \\
& & Spike            & $4.782 \pm 0.340$ & \textbf{10.0/10} & $\mathbf{0.458 \pm 0.015}$ \\
\midrule

\multirow{12}{*}{Same side}
& \multirow{6}{*}{Ant}
& Standard           & $4.060 \pm 0.332$ & \textbf{10.0/10} & $1.574 \pm 0.036$ \\
& & Pink             & $4.117 \pm 0.349$ & \textbf{10.0/10} & $1.231 \pm 0.034$ \\
& & Gaussian-Poisson & $4.111 \pm 0.392$ & \textbf{10.0/10} & $1.051 \pm 0.032$ \\
& & Poisson          & $\mathbf{4.370 \pm 0.350}$ & 9.2/10 & $0.785 \pm 0.029$ \\
& & Gaussian-Spike   & $4.207 \pm 0.364$ & \textbf{10.0/10} & $0.886 \pm 0.028$ \\
& & Spike            & $4.342 \pm 0.311$ & \textbf{10.0/10} & $\mathbf{0.605 \pm 0.021}$ \\
\cmidrule(lr){2-6}

& \multirow{6}{*}{Ant-2}
& Standard           & $3.953 \pm 0.319$ & \textbf{10.0/10} & $1.060 \pm 0.024$ \\
& & Pink             & $4.069 \pm 0.317$ & \textbf{10.0/10} & $0.847 \pm 0.022$ \\
& & Gaussian-Poisson & $\mathbf{4.418 \pm 0.311}$ & \textbf{10.0/10} & $0.663 \pm 0.019$ \\
& & Poisson          & $3.989 \pm 0.284$ & \textbf{10.0/10} & $0.562 \pm 0.018$ \\
& & Gaussian-Spike   & $4.381 \pm 0.358$ & \textbf{10.0/10} & $0.596 \pm 0.018$ \\
& & Spike            & $4.411 \pm 0.328$ & \textbf{10.0/10} & $\mathbf{0.476 \pm 0.014}$ \\
\bottomrule
\end{tabular}%
}
\end{table}

\begin{table}[!htbp]
\centering
\caption{Robustness to planner--plant integrator mismatch.}
\label{tab:integrator-mismatch}
\scriptsize
\setlength{\tabcolsep}{2.5pt}
\renewcommand{\arraystretch}{1.05}

\resizebox{\columnwidth}{!}{%
\begin{tabular}{lllccc}
\toprule
Mismatch & Robot & Sampler
& Vel. (m/s) $\uparrow$
& Laps/seed $\uparrow$
& Smooth. $\downarrow$ \\
\midrule

\multirow{12}{*}{Fast RK4}
& \multirow{6}{*}{Ant}
& Standard           & $4.584 \pm 0.347$ & 7.0/10 & $1.573 \pm 0.051$ \\
& & Pink             & $4.749 \pm 0.321$ & \textbf{8.9/10} & $1.243 \pm 0.045$ \\
& & Gaussian-Poisson & $4.782 \pm 0.339$ & 7.3/10 & $1.066 \pm 0.038$ \\
& & Poisson          & $4.651 \pm 0.368$ & 5.7/10 & $0.807 \pm 0.029$ \\
& & Gaussian-Spike   & $\mathbf{4.831 \pm 0.336}$ & 7.2/10 & $0.906 \pm 0.030$ \\
& & Spike            & $4.732 \pm 0.305$ & 6.9/10 & $\mathbf{0.624 \pm 0.022}$ \\
\cmidrule(lr){2-6}

& \multirow{6}{*}{Ant-2}
& Standard           & $4.343 \pm 0.296$ & \textbf{7.8/10} & $1.041 \pm 0.036$ \\
& & Pink             & $4.564 \pm 0.333$ & 6.3/10 & $0.831 \pm 0.027$ \\
& & Gaussian-Poisson & $4.841 \pm 0.331$ & 7.0/10 & $0.657 \pm 0.023$ \\
& & Poisson          & $4.224 \pm 0.323$ & 6.3/10 & $0.574 \pm 0.022$ \\
& & Gaussian-Spike   & $\mathbf{4.903 \pm 0.306}$ & 7.2/10 & $0.604 \pm 0.022$ \\
& & Spike            & $4.852 \pm 0.307$ & 5.0/10 & $\mathbf{0.484 \pm 0.017}$ \\
\midrule

\multirow{12}{*}{Implicit fast}
& \multirow{6}{*}{Ant}
& Standard           & $4.055 \pm 0.461$ & 9.1/10 & $1.587 \pm 0.051$ \\
& & Pink             & $4.142 \pm 0.496$ & \textbf{9.5/10} & $1.237 \pm 0.038$ \\
& & Gaussian-Poisson & $\mathbf{4.232 \pm 0.507}$ & \textbf{9.5/10} & $1.035 \pm 0.028$ \\
& & Poisson          & $4.009 \pm 0.451$ & 8.2/10 & $0.798 \pm 0.025$ \\
& & Gaussian-Spike   & $4.138 \pm 0.433$ & 9.0/10 & $0.882 \pm 0.023$ \\
& & Spike            & $4.012 \pm 0.420$ & 8.0/10 & $\mathbf{0.617 \pm 0.022}$ \\
\cmidrule(lr){2-6}

& \multirow{6}{*}{Ant-2}
& Standard           & $3.616 \pm 0.471$ & 8.8/10 & $1.131 \pm 0.051$ \\
& & Pink             & $3.666 \pm 0.421$ & 7.5/10 & $0.890 \pm 0.032$ \\
& & Gaussian-Poisson & $4.057 \pm 0.447$ & \textbf{9.4/10} & $0.691 \pm 0.032$ \\
& & Poisson          & $3.654 \pm 0.402$ & 8.0/10 & $0.582 \pm 0.022$ \\
& & Gaussian-Spike   & $\mathbf{4.214 \pm 0.529}$ & 8.5/10 & $0.612 \pm 0.025$ \\
& & Spike            & $3.963 \pm 0.411$ & 3.9/10 & $\mathbf{0.497 \pm 0.015}$ \\
\bottomrule
\end{tabular}%
}
\end{table}

\paragraph{Altered morphology}
All methods remain comparatively robust to altered leg-length configurations (Table~\ref{tab:leg-mismatch}), with almost all conditions completing the full evaluation, and the smoothness ordering observed in flat racing is preserved: Spike has the lowest control-increment magnitude in every condition ($p_\mathrm{H}=0.008$).

\paragraph{Integrator mismatch}
Integrator mismatch, however, exposes a clearer trade-off between task velocity, control smoothness, and completion reliability (Table~\ref{tab:integrator-mismatch}). Spike remains the smoothest method in all four conditions ($p_\mathrm{H}\leq 0.016$), while the highest mean velocity is attained by the spectrum-matched Gaussian controls in each of these scenarios. Moreover, Spike's smoothness does not translate into greater completion: on Ant-2, Spike completes only $5.0$ and $3.9$ laps/seed under the two mismatches, compared with $7.8$ and $8.8$ for Standard ($p_\mathrm{H}=1.0$ and $0.094$) and $7.2$ and $8.5$ for Gaussian-Spike ($p_\mathrm{H}=1.0$ and $0.094$). The integrator experiments therefore suggest that the temporal structure that yields smoother and, in several cases, faster locomotion can become less robust when the predictive model differs substantially from the plant.

\subsection{Reaching Task}

\begin{figure*}[!t]
    \centering
    \includegraphics[width=0.85\textwidth]{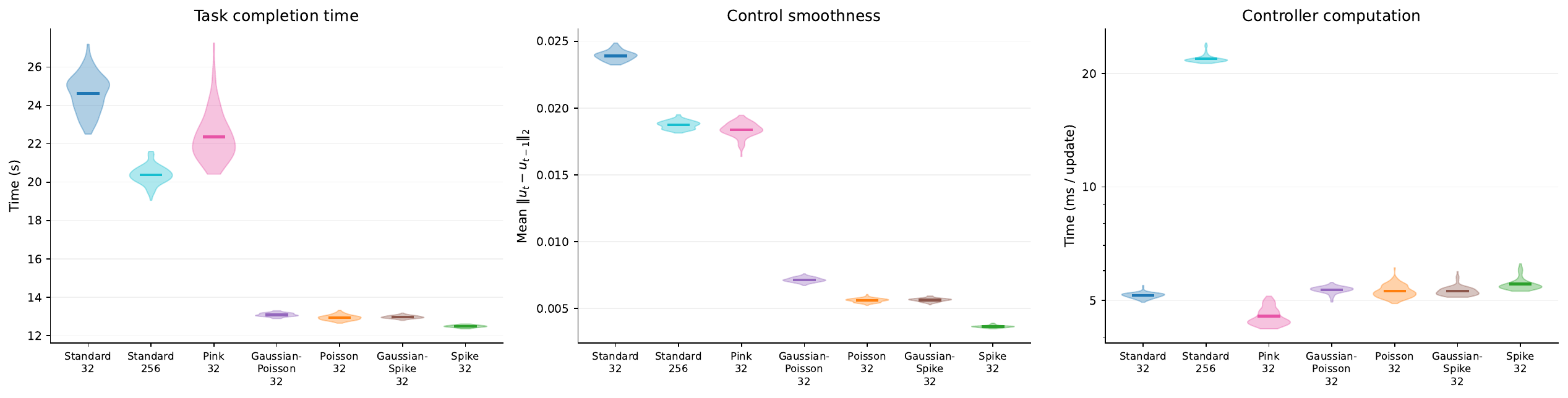}
    \caption{Reaching task performances for the FR3 robot. From left to right: completion time, control smoothness and controller computation. All metrics are shown over 100 runs and all samplers are evaluated with 32 rollouts except for the Standard sampler which is also evaluated with 256 rollouts.}
    \label{fig:reaching-violins}
\end{figure*}

Figure~\ref{fig:reaching-violins} compares the samplers on the reaching task. Standard Gaussian sampling with $N=256$ rollouts is included as a descriptive high-budget reference and is not part of the inferential comparisons. Every method completed the full target sequence in all 100 runs. All twelve hypothesis-driven contrasts are significant ($p\approx10^{-5}$, the resolution limit of the test).

Structured proposals substantially shorten task completion. Mean completion time decreases from $24.62~\mathrm{s}$ with Standard and $22.36~\mathrm{s}$ ($-9.2\%$) with Pink to $12.94~\mathrm{s}$ with Poisson and $12.49~\mathrm{s}$ with Spike, reductions of $47.4\%$ and $49.3\%$, respectively. Spike is also faster relative to Gaussian-Spike ($12.97~\mathrm{s}$). Notably, Spike with $N=32$ completes the task $38.7\%$ faster than Standard with $N=256$ ($20.38~\mathrm{s}$), at approximately one quarter of the computation. This advantage also arises with a reduced run-to-run variability, from a standard deviation of $0.96~\mathrm{s}$ with Standard to $0.06~\mathrm{s}$ with Spike.

As in locomotion, the strongest effect of the proposal structure is observed in executed-control smoothness. Relative to Standard, the mean control-increment norm $\|\mathbf u_t-\mathbf u_{t-1}\|_2$ is reduced by $23.2\%$ with Pink, $70.2\%$ with Gaussian-Poisson, $76.6\%$ with Poisson, $76.4\%$ with Gaussian-Spike, and $84.8\%$ with Spike. Increasing the Standard rollout budget to $N=256$ reduces it by only $21.6\%$. Spike produces the smoothest commands of all methods, with increments $35.0\%$ smaller than Poisson and $35.4\%$ smaller than Gaussian-Spike, indicating that the motor-pool structure contributes to smoothness beyond what is recovered by matching its power spectrum.

The structured proposals incur a modest computational overhead ($5.15$ to $5.53~\mathrm{ms/step}$ from Standard to Spike), small compared with the $21.89~\mathrm{ms/step}$ of Standard with $N=256$ and well within the $10~\mathrm{ms}$ budget of the $100~\mathrm{Hz}$ loop.

\subsection{Generalization to Manipulation Tasks}

\begin{table}[!htbp]
\centering
\caption{Ball balancing on the Franka plate. Error is the mean in-plane distance of the ball from the plate centre over each run.}
\label{tab:balancing}
\scriptsize
\setlength{\tabcolsep}{2.5pt}
\renewcommand{\arraystretch}{1.05}

\resizebox{\columnwidth}{!}{%
\begin{tabular}{lccc}
\toprule
Sampler
& Error (mm) $\downarrow$
& Success (\%) $\uparrow$
& Smooth. ($\times 10^{-3}$) $\downarrow$ \\
\midrule
Standard           & $40.41 \pm 8.66$ & 0  & $40.41 \pm 1.37$ \\
Pink               & $31.99 \pm 7.04$ & 0  & $36.03 \pm 1.80$ \\
Gaussian-Poisson   & $15.93 \pm 2.99$ & 2  & $18.88 \pm 1.20$ \\
Poisson            & $6.67 \pm 3.88$  & 81 & $6.50 \pm 1.52$ \\
Gaussian-Spike     & $10.27 \pm 2.38$ & 11 & $13.81 \pm 1.09$ \\
Spike              & $\mathbf{3.39 \pm 0.83}$ & \textbf{98} & $\mathbf{4.27 \pm 0.41}$ \\
\bottomrule
\end{tabular}%
}
\end{table}

\paragraph{Balancing a ball} Table~\ref{tab:balancing} reports the ball-balancing results. With $N=32$ rollouts, neither Standard nor Pink sampling succeeds in any of the 100 runs, with mean ball errors of $40.4$ and $32.0~\mathrm{mm}$. The structured proposals markedly improve performance: Poisson keeps the ball on the plate in $81\%$ of the runs, and Spike in $98\%$ ($p<10^{-3}$), with the lowest mean error ($3.39~\mathrm{mm}$, $49\%$ below Poisson; $p<10^{-4}$). In contrast, the spectrum-matched Gaussian controls succeed in only $2\%$ (Gaussian-Poisson) and $11\%$ (Gaussian-Spike) of the runs ($p<10^{-4}$ against Spike for both), indicating that the task-level benefit of the event-based proposals is not explained by their matched finite-horizon spectra alone. The reduced control increments provide one plausible mechanism. Consistent with the reaching results, Spike also produces the smoothest commands, reducing the mean control increment by $89\%$ relative to Standard and by $34\%$ relative to Poisson ($p<10^{-4}$ for both).

\begin{table}[!htbp]
\centering
\caption{Ball bouncing on a racket. Apex error is the mean absolute difference between each apex reached between two successive racket contacts and the $20~\mathrm{cm}$ target apex. A run succeeds when the ball reaches 50 bounces without being lost.}
\label{tab:bouncing}
\scriptsize
\setlength{\tabcolsep}{2.5pt}
\renewcommand{\arraystretch}{1.05}

\resizebox{\columnwidth}{!}{%
\begin{tabular}{lccc}
\toprule
Sampler
& Apex err. (mm) $\downarrow$
& Success (\%) $\uparrow$
& Smooth. ($\times 10^{-3}$) $\downarrow$ \\
\midrule
Standard           & $79.45 \pm 26.28$ & 64 & $51.14 \pm 0.95$ \\
Pink               & $85.95 \pm 39.60$ & 96 & $45.43 \pm 0.92$ \\
Gaussian-Poisson   & $69.85 \pm 44.02$ & \textbf{99} & $30.37 \pm 0.92$ \\
Poisson            & $57.66 \pm 29.47$ & 91 & $28.66 \pm 1.38$ \\
Gaussian-Spike     & $74.99 \pm 42.40$ & 92 & $23.96 \pm 0.80$ \\
Spike              & $\mathbf{55.32 \pm 34.42}$ & 90 & $\mathbf{20.06 \pm 0.84}$ \\
\bottomrule
\end{tabular}%
}
\end{table}

\paragraph{Bouncing a ball} Table~\ref{tab:bouncing} reports the ball-bouncing results. Unlike balancing, keeping the ball alive is achievable by all proposals: success ranges from $64\%$ with Standard to $99\%$ with Gaussian-Poisson, and Spike succeeds in $90\%$ of the runs. This rate is lower than that of Gaussian-Poisson ($p=0.012$), but not significantly different from Pink, Poisson, or Gaussian-Spike ($p\geq0.18$). However, Spike tracks the target height most accurately, with the lowest apex error ($55.3~\mathrm{mm}$), significantly lower than Standard ($79.5~\mathrm{mm}$), Pink ($86.0~\mathrm{mm}$), Gaussian-Poisson ($69.9~\mathrm{mm}$; $p=0.011$), and Gaussian-Spike ($75.0~\mathrm{mm}$; $p<10^{-3}$), and comparable to Poisson ($57.7~\mathrm{mm}$; $p=0.60$). All methods undershoot the target on average, but Spike comes closest, with a mean apex of $18.6~\mathrm{cm}$ compared with $15.0$--$15.8~\mathrm{cm}$ for Standard and the spectrum-matched controls. Since lower bounces are easier to sustain, the higher success rates of these baselines partly reflect a less demanding interpretation of the task. Spike again produces the smoothest commands, reducing the mean control increment by $61\%$ relative to Standard and by $30\%$ relative to Poisson ($p<10^{-4}$ for all comparisons).

\begin{table}[!htbp]
\centering
\caption{Box pushing on the Franka table. The box must be pushed through a three-waypoint route within $20~\mathrm{s}$, holding each waypoint within $2.5~\mathrm{cm}$ for $1~\mathrm{s}$. Task time is computed over successful runs only.}
\label{tab:pushing}
\scriptsize
\setlength{\tabcolsep}{2.5pt}
\renewcommand{\arraystretch}{1.05}

\resizebox{\columnwidth}{!}{%
\begin{tabular}{lccc}
\toprule
Sampler
& Task time (s) $\downarrow$
& Success (\%) $\uparrow$
& Smooth. ($\times 10^{-3}$) $\downarrow$ \\
\midrule
Standard           & --- & 0  & $44.51 \pm 5.76$ \\
Pink               & $12.68 \pm 2.74$ & 52 & $34.17 \pm 6.31$ \\
Gaussian-Poisson   & $10.59 \pm 2.62$ & 59 & $21.82 \pm 5.53$ \\
Poisson            & $11.26 \pm 2.74$ & 61 & $21.31 \pm 4.26$ \\
Gaussian-Spike     & $\mathbf{9.59 \pm 2.24}$ & 64 & $15.99 \pm 3.86$ \\
Spike              & $10.02 \pm 2.81$ & \textbf{75} & $\mathbf{13.82 \pm 2.88}$ \\
\bottomrule
\end{tabular}%
}
\end{table}

\paragraph{Pushing a box} Table~\ref{tab:pushing} reports the box-pushing results. Standard sampling fails in all 100 runs, while Spike achieves the highest success rate ($75\%$), significantly above Pink ($52\%$; $p<10^{-3}$), Gaussian-Poisson ($59\%$; $p=0.030$), and Poisson ($61\%$; $p=0.034$), and numerically above Gaussian-Spike ($64\%$; $p=0.12$). Among successful runs, Spike completes the route in $10.0~\mathrm{s}$, slightly longer than Gaussian-Spike ($9.6~\mathrm{s}$), although this difference is not significant on the seeds where both methods succeed ($+0.20~\mathrm{s}$, $p=0.72$). Spike remains faster than Poisson ($p=0.015$) and Pink ($p<10^{-3}$) on paired successful runs. Finally, Spike once more yields the smoothest commands, reducing the mean control increment by $69\%$ relative to Standard, $35\%$ relative to Poisson, and $14\%$ relative to Gaussian-Spike ($p<10^{-3}$ for all comparisons).

\begin{figure*}[!t]
    \centering
    \includegraphics[width=0.85\textwidth]{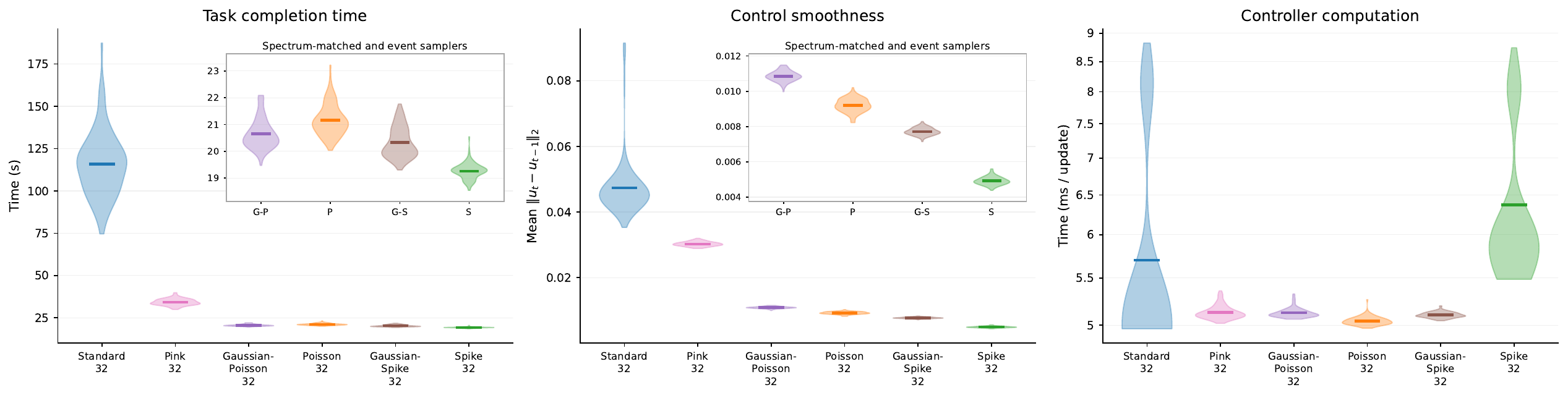}
    \caption{Reaching task performances for the sim-to-real FR3 robot. From left to right: completion time, control smoothness and controller computation. All metrics are shown over 100 runs and all samplers are evaluated with 32 rollouts.}
    \label{fig:reaching-real-violins}
\end{figure*}

\subsection{Reaching on a Real FR3 Robot}

Figure~\ref{fig:reaching-real-violins} compares the samplers on the physical FR3. All structured and colored proposals completed the full target sequence in every run, whereas Standard sampling failed in 7 runs (by drifting out of the workspace), a significantly lower success rate ($93\%$, $p=0.016$). All eight hypothesis-driven contrasts on completion time and control smoothness are significant ($p\approx10^{-5}$).

Structured proposals substantially shorten task completion on hardware. Mean completion time over successful runs decreases from $115.91~\mathrm{s}$ with Standard and $34.31~\mathrm{s}$ with Pink to $21.16~\mathrm{s}$ with Poisson and $19.26~\mathrm{s}$ with Spike. Spike is also faster than the spectrum-matched Gaussian-Spike ($20.33~\mathrm{s}$) and Gaussian-Poisson ($20.66~\mathrm{s}$) controls ($p\approx10^{-5}$ for both). This advantage comes with a markedly reduced run-to-run variability, from a standard deviation of $18.92~\mathrm{s}$ with Standard to $0.29~\mathrm{s}$ with Spike.

As in simulation, the strongest effect of the proposal structure is observed in executed-control smoothness. Relative to Standard, the mean control-increment norm $\|\mathbf u_t-\mathbf u_{t-1}\|_2$ is reduced by $36.1\%$ with Pink, $77.1\%$ with Gaussian-Poisson, $80.6\%$ with Poisson, $83.7\%$ with Gaussian-Spike, and $89.6\%$ with Spike. Spike produces the smoothest commands of all methods, with increments $46.5\%$ smaller than Poisson and $36.3\%$ smaller than Gaussian-Spike ($p\approx10^{-5}$), indicating that the motor-pool structure contributes to smoothness beyond what is recovered by matching its power spectrum, and that this benefit transfers to the physical robot.

Mean computation time ranges from $5.04~\mathrm{ms/step}$ (Poisson) to $6.37~\mathrm{ms/step}$ (Spike); Spike is thus moderately more expensive on hardware ($p\approx10^{-5}$), but all methods remain within the $10~\mathrm{ms}$ budget of the $100~\mathrm{Hz}$ control loop.

\section{Discussion}
\label{sec:discussion}

Across all experiments, the Spike proposal consistently yields the smoothest executed commands, while its effect on task performance depends on the setting. The following discussion identifies where this effect originates in the planar ablation and how it shapes behavior in locomotion, in manipulation, and on the physical robot.

\subsection{Ablation}

In the planar benchmark, removing the common drive produces the largest deterioration in executed-control smoothness, followed by twitch heterogeneity and, to a lesser extent, rate coding. Because this intervention also adds high-frequency power, it identifies an important contribution of shared input without isolating it from the accompanying spectral change. The controls clarify the advantage of Spike over the spectrum-matched Gaussian: it persists with a stationary prehistory and is not reproduced by causally filtered Gaussian noise, while a Gaussian with the full nonstationary covariance of Spike recovers most but not all of it. This advantage therefore stems mainly from nonstationary second-order structure, with a smaller but significant contribution of the non-Gaussian event structure. Renewal discharge has only a marginal effect, slightly roughening the commands, and is therefore not required in this setting. Overall, full Spike combines near-minimal executed-control roughness with high coverage and among the shortest times to goal.

\subsection{Locomotion}

In locomotion, the most consistent effect of the proposed sampling structure is the substantial increase in executed-control smoothness while preserving task performance, with Spike smoother than Poisson, itself smoother than Standard, across terrain, task, and morphology variations.

The spectrum-matched Gaussian controls indicate that second-order temporal structure accounts for a substantial part of the task-level behavior: no significant Spike--Gaussian-Spike difference is detected in flat-running velocity or weighted rollout cost. The full Spike proposal nevertheless differs from Gaussian-Spike in executed control, with significantly different smoothness in all flat-running conditions and distinct co-contraction on Ant-2 across all rollout budgets. Because Gaussian-Spike matches the estimated finite-horizon second-order spectrum but not all nonstationary covariance structure, these residual effects are consistent with additional nonstationary dependencies and higher-order temporal or amplitude statistics associated with the motor-pool construction; the planar ablation indicates that the former account for most of the difference. Relative to Poisson, Spike differs mainly in smoothness and coordination, with task-level differences appearing primarily on Ant-2, suggesting that the benefit of the richer motor-pool structure depends on the actuation model.

\subsection{Manipulation}

The manipulation experiments confirm the smoothness gains observed in locomotion, with Spike yielding the lowest control increments on all four FR3 tasks. Unlike locomotion, however, these gains are accompanied by clear improvements in task performance, most pronounced on tasks that inherently require smooth behavior. On ball balancing, where abrupt plate motions destabilize the ball, Spike succeeds in almost all runs, whereas Standard, Pink, and the spectrum-matched controls almost always fail; contrary to locomotion, this benefit is not explained by the power spectrum alone. A similar pattern appears on box pushing and, for apex accuracy, on ball bouncing.

These results, together with the stronger locomotion differences at $N=32$, indicate that, at low rollout budgets, smoothness of the proposal is itself a desirable property. The MPPI update is a weighted average of the sampled perturbations. With many rollouts, high-frequency components that are not consistently associated with low cost tend to cancel in this average; with $N=32$, the update is dominated by a small number of low-cost samples whose high-frequency content passes largely unattenuated into the executed command. Accordingly, increasing the Standard budget from $N=32$ to $N=256$ reduces control increments on reaching by only $21.6\%$, whereas Spike reduces them by $84.8\%$ with one eighth of the rollouts. On tasks whose dynamics are sensitive to high-frequency actuation, such as balancing an object or maintaining contact while pushing, this residual jitter can separate success from failure.

\subsection{Sim-to-Real Transfer}

The hardware experiments show that the proposed samplers transfer to a physical FR3, but that transfer differs markedly between samplers. Standard degrades strongly: its mean completion time increases from $24.6~\mathrm{s}$ in simulation to $115.9~\mathrm{s}$, and $7$ of the $100$ runs had to be aborted. The command delay of the communication and joint-level tracking loop, together with the unmodeled dynamics of the real robot, amplify its jerky commands into visible shaking of the arm, which in turn perturbs the measured state used to initialize subsequent rollouts. Pink is also affected ($22.4$ to $34.3~\mathrm{s}$), whereas the structured proposals degrade only moderately and complete every run.

On hardware, Spike retains the shortest completion time and a smoothness advantage over Poisson and Gaussian-Spike at least as large as in simulation, suggesting that the smoothness of the proposal may contribute to robustness against delay and model--plant discrepancies.

\subsection{Conclusions, Limitations, and Future Work}

The experiments show that the temporal structure of the MPPI proposal substantially influences closed-loop behavior under finite sampling budgets, with three distinct findings. First, Spike consistently yields the smoothest executed commands, including relative to spectrum-matched Gaussian proposals, and this advantage transfers to a physical robot. Second, Spike improves task performance in several important scenarios, notably reaching in simulation and on hardware, ball balancing, and box pushing. Third, Spike does not consistently dominate the other proposals in task performance or reliability: Standard is faster on Ant with the full rollout budget, Gaussian-Poisson succeeds more often in ball bouncing, and Spike completes fewer laps under integrator mismatch on Ant-2. These results support treating proposal design as a combination of spectral shaping and additional distributional or nonstationary structure when designing efficient, structured MPPI sampling distributions. The co-contraction and coordination differences between Spike and Poisson are compatible with a role for the common stochastic drive, which couples the agonist and antagonist pools of each coordinate, whereas Poisson uses independent positive and negative event streams. Consistent with this, removing the common drive degrades smoothness most among the planar ablations, although this intervention also alters the proposal spectrum and Spike and Poisson also differ in recruitment, discharge, and twitch dynamics \cite{henneman1957relation,fuglevand1993models,farina2014effective,hug2023common,caillet2023motoneuron}.

Several limitations remain. First, the antagonistic actuator and motoneuron models are simplified abstractions rather than quantitatively validated muscle--tendon or physiological models. Second, the structured samplers introduce a modest computational overhead. Third, the integrator-mismatch results show that a stronger temporal proposal prior is not uniformly beneficial under substantial model error. This sim-to-sim sensitivity contrasts with the successful sim-to-real transfer on the FR3 and warrants further study of how the temporal scales of the proposal interact with model error. Finally, the hardware evaluation is limited to reaching; more complex tasks, such as pushing, balancing, or bouncing, as well as locomotion, remain to be validated on physical robots.

The event-based formulation also creates several directions for future work. Because the latent perturbation is represented by sparse event times and marks, optimization could operate directly over event sequences, for example through cross-entropy search with Bernoulli or categorical distributions over event occurrence, sign, and recruitment depth \cite{de2005tutorial,rubinstein1999cross}. A second direction is to learn proposal parameters, such as event rates, temporal scales, recruitment structure, or actuator synergies, from control performance. MPPI could also act as a teacher for learning a closed-loop policy, following approaches in guided policy search \cite{levine2013guided, ross2011reduction, kahn2017plato, reske2021imitation}. More broadly, Spike sampling is not specific to MPPI and could be evaluated as an exploration process in reinforcement learning, where colored action noise has already been shown to improve exploration \cite{eberhard2023pink}, both in off-policy methods and as a structured action distribution in on-policy methods such as PPO \cite{schulman2017proximal}.

\hypersetup{urlcolor = black}

\bibliographystyle{IEEEtran}
\bibliography{biblio}

\end{document}